\documentclass[10pt,twocolumn,letterpaper]{article}

\usepackage{wacv}
\usepackage{times}
\usepackage{epsfig}
\usepackage{graphicx}
\usepackage{amsmath}
\usepackage{amssymb}

\usepackage{xcolor}
\usepackage{soul}
\usepackage[utf8]{inputenc}
\usepackage[pagebackref=true,breaklinks=true,letterpaper=true,colorlinks,bookmarks=false]{hyperref}

\usepackage[figuresright]{rotating}
\usepackage{textcomp,subfigure,algorithm,algorithmic,multirow,upgreek,tabularx}
\usepackage{graphics,gensymb}
\usepackage{bm,cases,threeparttable}
\def\R{\mathbb{R}}

\usepackage{setspace}

\usepackage{enumitem}
\usepackage{caption}
\setitemize{noitemsep,topsep=0pt,parsep=0pt,partopsep=0pt}
\setenumerate{noitemsep,topsep=0pt,parsep=0pt,partopsep=0pt}

\wacvfinalcopy % *** Uncomment this line for the final submission

\def\wacvPaperID{****} % *** Enter the wacv Paper ID here

\ifwacvfinal\fi
\begin{document}

%%%%%%%%% TITLE
\title{Deep Micro-Dictionary Learning and Coding Network}

% Authors at the same institution
%\author{First Author \hspace{2cm} Second Author \\
% Authors at different institutions

%\author{First Author \\
%\and
%Second Author \\

\author{Hao Tang$^1$, Heng Wei$^2$, Wei Xiao$^3$, Wei Wang$^{4}$, Dan Xu$^{5}$, Yan Yan$^{6}$, Nicu Sebe$^1$\\
$^1$Department of Information Engineering and Computer Science, University of Trento, Trento, Italy \\
$^2$Department of Electrical Engineering, Hong Kong Polytechnic University, Hong Kong, China\\
$^3$Lingxi Artificial Intelligence Co., Ltd, Shen Zhen, China\\
$^4$Computer Vision Laboratory, \'Ecole Polytechnique F\'ed\'erale de Lausanne, Lausanne, Switzerland \\
$^5$Department of Engineering Science, University of Oxford, Oxford, UK \\
$^6$Department of Computer Science, Texas State University, San Marcos, USA \\
{\tt\small \{hao.tang, niculae.sebe\}@unitn.it, 15102924d@connect.polyu.hk, xiaoweithu@163.com}\\
{\tt\small wei.wang@epfl.ch, danxu@robots.ox.ac.uk, y\_y34@txstate.edu}\\
}

\maketitle
\ifwacvfinal\thispagestyle{empty}\fi

\begin{abstract}
	In this paper, we propose a novel \textbf{D}eep Micro-\textbf{D}ictionary \textbf{L}earning and \textbf{C}oding \textbf{N}etwork (\textbf{DDLCN}).
	DDLCN has most of the standard deep learning layers (pooling, fully, connected, input/output, etc.) but the main difference is that the fundamental convolutional layers are replaced by novel compound dictionary learning and coding layers.  
	The dictionary learning layer learns an over-complete dictionary for the input training data.
	At the deep coding layer, a locality constraint is added to guarantee that the activated dictionary bases are close to each other. 
	Next, the activated dictionary atoms are assembled together and passed to the next compound dictionary learning and coding layers.
	In this way, the activated atoms in the first layer can be represented by the deeper atoms in the second dictionary.
	Intuitively, the second dictionary is designed to learn the fine-grained components which are shared among the input dictionary atoms.
	In this way, a more informative and discriminative low-level representation of the dictionary atoms can be obtained. 
	We empirically compare the proposed DDLCN with several dictionary learning methods and deep learning architectures. 
	The experimental results on four popular benchmark datasets demonstrate that the proposed DDLCN achieves competitive results compared with state-of-the-art approaches.
\end{abstract}

\begin{figure}[!h] \tiny
	\begin{centering}
		\includegraphics[width=1\linewidth]{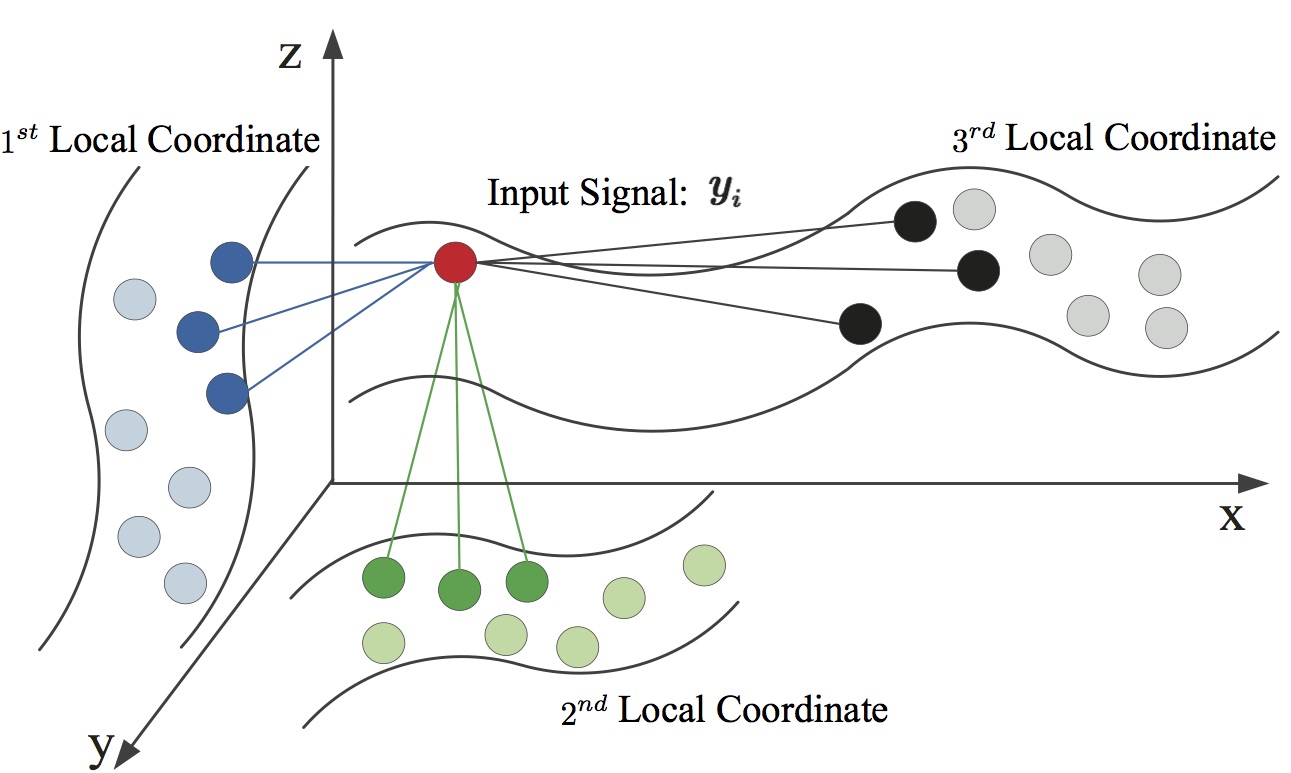}\\
		\caption{Multiple local coordinates and ``fake" anchor points.
			Suppose that we want to encode the input signal $\bm{y}_i{\in} \R^m$ in order to find the nonlinear function defined on it, for instance with three local coordinates in the same space $\R^m$.
			In fact, $\bm{y}_i$ relies on the manifold that can be described by the $3^{rd}$ local coordinate, but unfortunately, we do not have enough atoms (anchor points) on the  $3^{rd}$ manifold close to it.
			Therefore, some nearby atoms (``fake'' anchor
			points) from the $1^{st}$ or $2^{nd}$ local coordinate will ``kidnap'' $\bm{y}_i$, overlooking in this way the true coordinate ($3^{rd}$), where
			$\bm{y}_i$ really resides. This happens due to the small numbers of training samples and/or a large variety of signals, arriving at overfitting
			even though the learned model fits the training samples well.
		}
		\label{manifold}
	\end{centering}
	\vspace{-0.4cm}
\end{figure}

\begin{figure*}[!tbp]\tiny
	\begin{center}
		\includegraphics[width=1\linewidth]{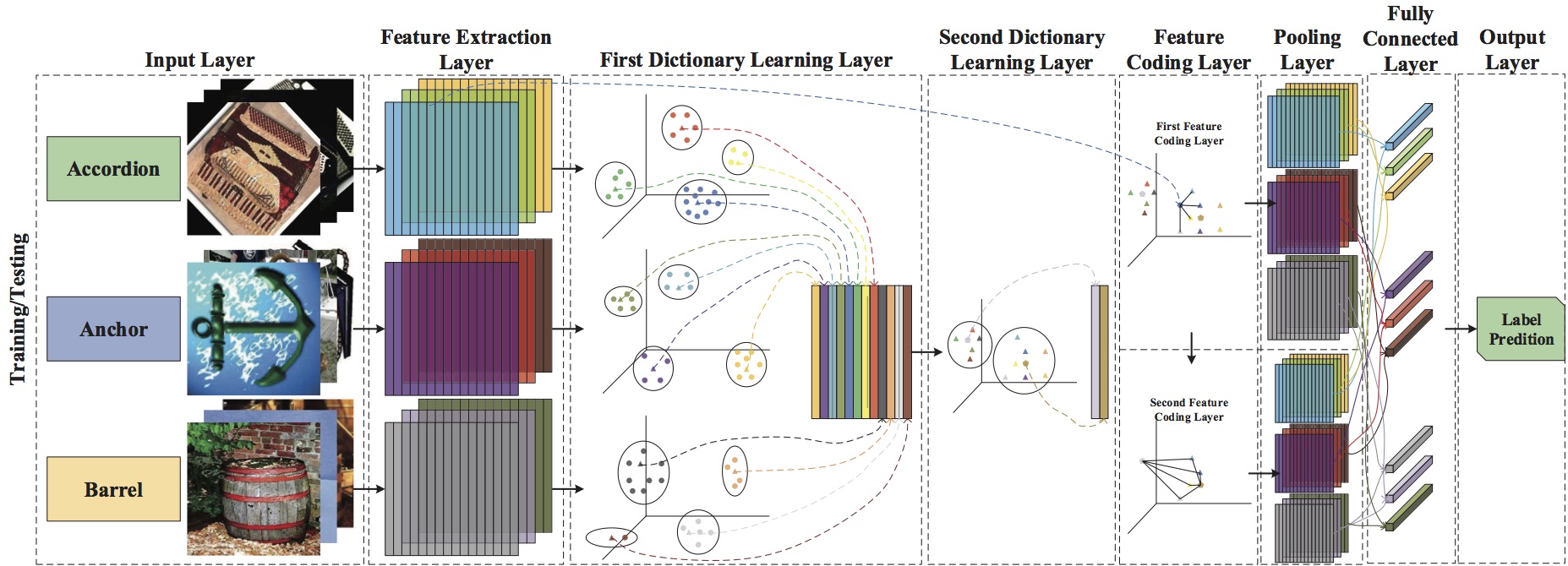}\\
		\caption{The pipeline of the Deep Micro-Dictionary Learning and Coding Network (DDLCN).
			The idea of DDLCN comes from the architectures of CNNs, while the difference from CNNs is that the convolutional layers in CNNs are replaced by our compound dictionary learning and coding layers.}
		\label{fig:framework}
	\end{center}
	\vspace{-0.8cm}
\end{figure*}

\section{Introduction}
In the past few years, the most popular representation learning frameworks are dictionary learning and deep learning.
Dictionary learning aims at learning a set of atoms such that a given feature can be well approximated by a sparse linear combination of these atoms, while deep learning methods focus on extracting semantic features via a deep network. 
%while deep learning methods focus on extracting features via network weights between the input and the hidden layer. 
So far most studies in dictionary learning employ a shallow (single layer) architecture, \emph{e.g.}, currently popular dictionary learning techniques are  K-SVD \cite{aharon2006k}, Discriminative K-SVD (D-KSVD) \cite{zhang2010discriminative} and Label Consistent K-SVD (LC-KSVD) \cite{jiang2011learning} which decompose the training data into a dense basis and sparse coefficients.
In addition, both Local Coordinate Coding (LCC) \cite{yu2009nonlinear,yu2010improved} and its fast implementation algorithm \cite{wang2010locality} are traditional dictionary learning methods.
LCC and Locality Constrained Coding (LLC) \cite{wang2010locality} are based on the empirical observation that the sparse representations tend to be ``local''.
In other words, nonzero coefficients are often assigned to the atoms nearby to the encoded signal $\bm{y}$.
However, LLC has a major disadvantage: to achieve higher approximation, one has to use a large number of so-called ``anchor points'' to make a better linear approximation of the signal.
Since LLC is a local linear approximation of a complex signal $\bm{y}_i$, for a nonlinear function on $\bm{y}_i$ the local linear approximation may not necessarily be optimal.
It means that the anchor points need to provide higher approximation power, allowing some of them to not necessary be ``real'' local anchors on the manifold where $\bm{y}_i$ resides.
In this context, our goal is to equip anchors with more descriptive power for better approximating $\bm{y}_i$ in order to finally make more accurate inferences from it.
An illustrative example is shown in Figure~\ref{manifold}.

Recent work \cite{tariyal2016greedy} has shown that deeper architectures can be built from dictionary learning. 
Chun et al.~\cite{chun2018convolutional} present a Block Proximal Gradient method using a Majorizer for convolutional dictionary learning.
Hu et al.~\cite{hu2018nonlinear} propose a nonlinear dictionary learning method and apply it to image classification task.
Xiao et al.~\cite{xiao2015two} propose a two-layer local coordinate coding framework for object recognition task.
Zhang et al.~\cite{zhang2017jointly} introduce an analysis discriminative dictionary learning framework for image classification task.
Nguyen et al.~\cite{nguyen2015dash} propose a domain adaptation framework using a sparse and hierarchical network, which shares the ideas with our work.
However, our DDLCN is different from~\cite{nguyen2015dash} in two ways:
(i) Our dictionary is learned from features and then the learned dictionary acts as a candidate pool for the next layer dictionary. 
Our dictionaries from different layers have connections while in \cite{nguyen2015dash} which used a fixed dictionary in different layers, \textit{i.e.}, there is no message passing between the dictionaries of different layers;
(ii) To represent an atom in the previous layer, we pick out a few atoms in the next layer and linearly combine them. 
These atoms have a linear contribution in constructing the atom in the previous layer. 
This is vital for the diversity, and in this way could incorporate more information into the next layer’s codes and alleviate the influence of incorrect atoms. 
However, there is no such mechanism in~\cite{nguyen2015dash}.

Inspired by both dictionary and deep learning, the goal of this paper is to improve the deep representation ability of dictionary learning.
To address this problem, we present a novel network, named Deep Micro-Dictionary Learning and Coding Network (DDLCN), which is composed of several layers: input, feature extraction, dictionary learning, feature coding, pooling, fully connected and output layer as shown in Figure \ref{fig:framework}.
The idea of the DDLCN comes from the standard architecture of Convolutional Neural Networks (CNNs), the biggest difference being that the convolutional layers in CNNs are replaced by our compound dictionary learning and coding layers.
In this way, edges, lines and corners can be learned from the shallow layers which correspond to the shallow dictionaries. 
The more complicated ``hierarchical'' patterns/features can be obtained from deeper dictionaries.

DDLCN takes advantage of the manifold geometric structure of the underlying data to locally embed points from the underlying data manifold into a lower dimensional deep structural space.
The benefit of DDLCN is that the learned feature representation after the feature learning and coding layers has a  better approximation capability of the original data, in other words, the deep dictionary learning structure can fully exploit the space where the data reside.
Meanwhile, the deep dictionary structure is very flexible, making it possible to use a micro dictionary, \emph{e.g.}, we can learn only one dictionary item per category.

Our contributions are summarized as follows: 
\begin{itemize}[leftmargin=*]
	\item We propose a novel compound dictionary learning and coding layer, which has the similar function as the convolutional layer in the standard deep learning architecture.
	\item We present a new deep dictionary learning framework named Deep Micro-Dictionary Learning and Coding Network (DDLCN), which combines the advantages of dictionary and deep learning methods.
	\item Exhaustive experiments on a broader range of datasets have been conducted, demonstrating that the proposed layer and framework outperform the existing dictionary learning methods and achieve competitive results compared with deep learning approaches.
\end{itemize}

\section{The Proposed DDLCN}
In this section, we sequentially introduce each layer of DDLCN.
For simplicity, we provide details on two layers of dictionary learning and coding of the DDLCN framework.
Extension of DDLCN to multiple layers is straight forward.

\noindent \textbf{Feature Extraction Layer.} 
Let $\bm{Y}$ denote a set of $m$-dimensional local descriptors, which is extracted from the data sampled from some uni-modal sensors, \emph{i.e.}, $\bm{Y} {=} \left[ {{\bm{y}_1}, \cdots ,{\bm{y}_N}} \right] {\in} {\R^{m \times N}}$, where $N$ is the total number of local descriptors.
In order to emphasize the viability of the proposed deep dictionary learning and coding method, we only use a single descriptor, the Scale-Invariant Feature Transform (SIFT) \cite{lowe2004distinctive} throughout our experiment.
SIFT features used in dictionary learning are pretty common in the computer vision field \cite{shen2015multi,yan2015complex,yang2017top,kim2017modality,bao2016dictionary,zhu2015deep}. 
In our experiments, we achieve better performance using SIFT than using raw pixels.
For image $I$ in the dataset, we extract the SIFT feature $\bm{y_i}$ as:
\vspace{-0.2cm}
\begin{equation}
\bm{y}_i = F(I), i \in [1, ..., N],
\vspace{-0.2cm}
\end{equation}
where $F$ denotes the feature extractor.

\noindent \textbf{First Dictionary Learning Layer.} 
We assume the number of classes in the dataset is $r$ ($r {=}3$ in Figure~\ref{fig:framework}).
For each class, we select $p$ images to train the corresponding dictionary of the class, the size of this class denotes as $q$, as shown in Figure~\ref{fig:framework}, $p {=} 3$ and $q {=} 4$.
We adopted the following dictionary learning algorithm:
\vspace{-0.2cm}
\begin{equation}
\begin{array}{c}
\min\limits_{\bm{V}} \left[ \frac{1}{2} ||\bm{y}_i - \bm{V}\alpha_i||_2^2 \right]
\quad s.t.\quad ||\alpha_i||_1 <= \lambda
\end{array}
\vspace{-0.2cm}
\end{equation}
where $\bm{y}_i$ is the SIFT feature learned from $F$ and we set $\lambda {= }0.35$ in the following experiments.  
After learning the dictionary of each class, we group all dictionaries of each class to form the first dictionary $\bm{V} {=} \left[ {{\bm{v}_1},{\bm{v}_2}, \cdots ,{\bm{v}_{{s_1}}}} \right] \in {\R^{m \times {s_1}}}$.
Then the first dictionary learning layer contains $s_1$ entries and we have $s_1 {=} r {\times} q$ ($s_1 {=}3 {\times} 4 {=} 12$ in Figure~\ref{fig:framework}).

\noindent \textbf{Second Dictionary Learning Layer.} 
The second layer dictionary $\bm{U} {=} \left[ {{\bm{u}_1},{\bm{u}_2}, \cdots ,{\bm{u}_{{s_2}}}} \right]$ is obtained by learning from  the first layer codebook $\bm{V}$.
\vspace{-0.2cm}
\begin{equation}
\begin{array}{c}
\min\limits_{\bm{U}} \left[\frac{1}{2} ||\bm{v}_i - \bm{U}\alpha_i||_2^2 \right]
\quad s.t.\quad ||\alpha_i||_1 <= \lambda
\end{array}
\vspace{-0.2cm}
\end{equation}
where ${\bm{v}_i}\in\bm{V}$ is one of the basis vectors in the first dictionary ($\lambda {=} 0.35$), as shown in Figure~\ref{fig:framework}, $s_2 {=} 2$.

\noindent \textbf{The $\text{n}^{th}$ Dictionary Learning Layer.} 
We learn the $\text{n}^{th}$ dictionary $\bm{D^{n}} {=} \left[ {{\bm{d}_1^{n}},{\bm{d}_2^{n}}, \cdots ,{\bm{d}_{{s_n}}^{n}}} \right]$ from the previous layer dictionary $\bm{D^{n-1}}$.
\vspace{-0.2cm}
\begin{equation}
\begin{array}{c}
\min\limits_{\bm{D^{n}}} \left[ \frac{1}{2} ||\bm{d}_i^{n-1} - \bm{D^{n}}\alpha_i||_2^2 \right]
\quad s.t.\quad ||\alpha_i||_1 <= \lambda,
\end{array}
\vspace{-0.2cm}
\end{equation}
where ${\bm{d}_i^{n-1}}{\in} \bm{D^{n-1}}$ is one of the basis vectors in the $\text{(n-1)}^{th}$ dictionary layer.   

\noindent \textbf{First Feature Coding Layer.}
After obtaining the dictionary $\bm{V}$, each feature is then encoded by $\bm{V}$ through several nearest items to produce the first coding.
The number of the nearest items of the first coding layer is set to a small value (\emph{e.g.}, 15).
The first feature coding scheme converts each local descriptor $\bm{y}_i$ into a $s_1$ dimensional code $\bm{\gamma} _i^1 {=} {\left[ {\bm{\gamma} _i^1({\bm{v}_1}),\bm{\gamma} _i^1({\bm{v}_2}), \cdots ,\bm{\gamma} _i^1({\bm{v}_{{s_1}}})} \right]^\mathsf{T}} \in {\R^{s_1}}$.
We arrange each code corresponding to each descriptor into a matrix: ${\bm{\gamma} ^1} {=} \left[ {\bm{\gamma} _1^1,\bm{\gamma} _2^1, \cdots ,\bm{\gamma} _N^1} \right] \in {\R^{{s_1} \times N}}$.
Specifically, each code can be obtained using the following optimization:
\vspace{-0.2cm}
\begin{equation}
\label{first_layer_formulation}
\begin{array}{c}
\mathop {\min }\limits_{\bm{\gamma} _i^1} \left[ {\frac{1}{2}\left\| {{\bm{y}_i} - \bm{V}\bm{\gamma} _i^1} \right\|_2^2 + \beta {{\left\| {\bm{\gamma} _i^1 \odot {\bm{\zeta}_i^1}} \right\|}_1}} \right] \\
\quad s.t.\quad {\bm{1}^\mathsf{T}}\bm{\gamma} _i^1 = 1,
\end{array}
\vspace{-0.2cm}
\end{equation}
where $\bm{\zeta}_i^1 {\in} \R^{s_1}$ is a distance vector, measuring the distance between $\bm{y}_i$ and $\bm{v}_i$, and $ \odot $ denotes the element-wise multiplication or Hadamard product, which enables the corresponding items of both vectors ($\bm{\gamma} _i^1$  and $\bm{\zeta}_i^1$) to multiply.
Typically, $\bm{\zeta}_i^1$ can be obtained using $\ell_2$ norm, that is ${\bm{\zeta}_i^1} {=} \left[ {{{\left\| {{\bm{y}_i} - {\bm{v}_1}} \right\|}_2},{{\left\| {{\bm{y}_i} - {\bm{v}_2}} \right\|}_2}, \cdots ,{{\left\| {{\bm{y}_i} - {\bm{v}_{{s_1}}}} \right\|}_2}} \right]^\mathsf{T}$.

\begin{figure}[!t] \tiny
	\begin{center}
		\includegraphics[width=1\linewidth]{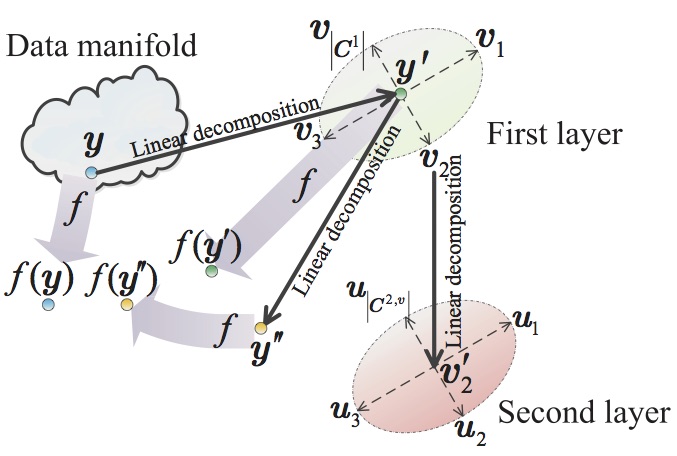}\\
		\caption{Multi layers coding strategy. 
			$\bm{y}$ is linearly combined with $\bm{v_1}$, $\bm{v_2}$, $\bm{v_3}$ and $\bm{v_{\left| {{\bm{C}^1}} \right|}$}, this is what we called single layer coding.
			Further, the atoms in the first layer can be represented by the atoms in the second layer.
			In other words, $\bm{v_1}$, $\bm{v_2}$, $\bm{v_3}$ and $\bm{v_{\left| {{\bm{C}^1}} \right|}$} are linearly represented by the atoms in the second layer, respectively.
			For example, one of these atoms $\bm{v_2}$, is combined by $\bm{u_1}$, $\bm{u_2}$, $\bm{u_3}$ and $\bm{u_{\left| {{\bm{C}^{2,\bm{v}}}} \right|}$} linearly.
			This is what we called two layers coding.
			By that analogy, we can obtain $n^{th}$ layer' codes.}
		\label{two_layer}
	\end{center}
	\vspace{-0.4cm}
\end{figure}

\noindent \textbf{Second Feature Coding Layer.}
Similarly, at the second layer, we have, 
\vspace{-0.2cm}
\begin{equation}
\label{formulation_second_layer}
\begin{array}{c}
\min\limits_{\bm{\gamma} _i^{2}} \left[ {\frac{1}{2}\left\| {{\hat{\bm{v}}_i} - \bm{U}\bm{\gamma} _i^{2}} \right\|_2^2 + \beta {{\left\| {\bm{\gamma} _i^{2} \odot \bm{\zeta}_i^2} \right\|}_1}} \right] \\
\quad s.t.\quad {1^\mathsf{T}}\bm{\gamma} _i^{2} = 1,
\end{array}
\vspace{-0.2cm}
\end{equation}
where $\bm{\gamma} _i^{2} {=} {\left[ {\bm{\gamma} _i^{2}({\bm{u}_1}),\bm{\gamma} _i^{2}({\bm{u}_2}), \cdots ,\bm{\gamma} _i^{2}({\bm{u}_{{s_2}}})} \right]^\mathsf{T}} \linebreak[3] {\in} {\R^{{s_2}}}$ is the second coding and 
$\bm{\zeta}_i^2 {\in} {\R^{{s_2}}}$ is used to measure the distance between $\hat{\bm{v}}_i$ and each atom in the dictionary matrix $\bm{U}$.
$\hat{\bm{v}}_i {\in} \bm{V}$ is one of the basis vectors adopted in the representation of $\bm{y}_i$ at the first layer.
We decompose these nearest atoms $\hat{\bm{v}}_i$ in the first layer to acquire the second layer coding using the second layer dictionary.
For better understanding, there is an illustrative explanation in Figure~\ref{two_layer}.

\noindent \textbf{The $\text{n}^{th}$ Feature Coding Layer.}
We generalize our two layers framework to a deeper one,
\vspace{-0.2cm}
\begin{equation}
\begin{array}{c}
\min\limits_{\bm{\gamma} _i^{n}} \left[ {\frac{1}{2}\left\| {\hat{\bm{d}}_i^{n-1} - \bm{D^n}\bm{\gamma}_i^{n}} \right\|_2^2 + \beta {{\left\| {\bm{\gamma} _i^{n} \odot \bm{\zeta}_i^n} \right\|}_1}} \right] \\
\quad s.t.\quad {1^\mathsf{T}}\bm{\gamma}_i^{n} = 1,
\end{array}
\vspace{-0.2cm}
\end{equation}
where $\bm{\gamma} _i^{n}$ is the $\text{n}^{th}$ layer coding and 
$\bm{\zeta}_i^n$ is used to measure the distance between $\hat{\bm{d}}_i^{n-1}$ and each atom in the $\text{n}^{th}$ dictionary $\bm{D^n}$.
$\hat{\bm{d}}_i^{n-1} {\in} \bm{D^{n-1}}$ is one of the basis vectors adopted in the representation of $\bm{y}_i$ at the $\text{(n-1)}^{th}$ coding layer.

\noindent \textbf{Pooling Layer.}
After the feature coding layer, the pooling layer then takes over. 
For each image, we employ $1{\times}1$, $2{\times}2$ and $4{\times}4$  spatial pyramids matching with max-pooling.

\noindent \textbf{Fully Connected Layer.}
The final output of $n^{th}$ layer's codes for $\bm{y}_i$ can be obtained by integrating each layers' codes. 
More specifically, each item in the first layer's codes $\bm{\gamma} _i^1$ can be augmented into a vector, for instance, the $j^{th}$ item can be augmented into the form of ${\left[ {\bm{\gamma} _i^1({\bm{v}_j}),\;\bm{\gamma} _i^1({\bm{v}_j})[\bm{\gamma} _j^{2}({\bm{u}_1}),\bm{\gamma} _j^{2}({\bm{u}_2}), \cdots ,\bm{\gamma} _j^{2}({\bm{u}_{{s_2}}})]} \right]^\mathsf{T}}$ (please refer to Algorithm~\ref{alg:SA}).

\noindent \textbf{Output Layer.}
The Support Vector Machine (SVM) is adopted as the classifier.
The implementation of multiclass SVM is provided by LIBSVM \cite{chang2011libsvm}.

\begin{algorithm}[!t] \small
	\caption{The two-layer model of the DDLCN.}
	\label{alg:SA}
	\begin{algorithmic}[1]
		\REQUIRE:
		$\bm{Y} \in {\R^{m \times N}}$
		\ENSURE:
		$\bm{\gamma} _i$\\
		\STATE First dictionary learning: $ \quad\bm{V}\leftarrow\bm{V}_{Dictionary}$
		\STATE First locality constraint calculating:\\
		${\quad\bm{\zeta}_i^1} = \left[ {{{\left\| {{\bm{y}_i} - {\bm{v}_1}} \right\|}_2},{{\left\| {{\bm{y}_i} - {\bm{v}_2}} \right\|}_2}, \cdots ,{{\left\| {{\bm{y}_i} - {\bm{v}_{{s_1}}}} \right\|}_2}} \right]^\mathsf{T}$
		\STATE First feature coding:\\
		$\quad \textbf{for}\quad i=1\;to\;N$\\
		$\quad\begin{array}{l}
		\bm{\gamma} _i^1 \leftarrow \mathop {\min }\limits_{\bm{\gamma} _i^1} \left[ {\frac{1}{2}\left\| {{\bm{y}_i} - \bm{V}\bm{\gamma} _i^1} \right\|_2^2 + \beta {{\left\| {\bm{\gamma} _i^1 \odot {\bm{\zeta}_i^1}} \right\|}_1}} \right] \\
		\quad \quad s.t.\quad {1^\mathsf{T}}\bm{\gamma} _i^1 = 1
		\end{array}$\\
		$\quad \textbf{end}$
		
		\STATE Second dictionary learning: $\quad\bm{U}\leftarrow\bm{U}_{Dictionary}$
		\STATE Second locality constraint calculating:\\
		$\quad{\bm{\zeta}_i^2} = \left[ {{{\left\| {{\bm{v}_i} - {\bm{u}_1}} \right\|}_2},{{\left\| {{\bm{v}_i} - {\bm{u}_2}} \right\|}_2}, \cdots ,{{\left\| {{\bm{v}_i} - {\bm{u}_{{s_2}}}} \right\|}_2}} \right]^\mathsf{T}$
		\STATE Second feature coding:\\
		$\quad \textbf{for}\quad i=1\;to\;N$\\
		$\quad \begin{array}{l}
		\bm{\gamma} _i^{2} \leftarrow \mathop {\min }\limits_{\bm{\gamma} _i^{2}} \left[ {\frac{1}{2}\left\| {{\bm{v}_i} - \bm{U}\bm{\gamma} _i^{2}} \right\|_2^2 + \beta {{\left\| {\bm{\gamma} _i^{2} \odot \bm{\zeta}_i^2} \right\|}_1}} \right] \\
		\quad \quad s.t.\quad {1^\mathsf{T}}\bm{\gamma} _i^{2} = 1
		\end{array}$\\
		$\quad \textbf{end}$	
		\STATE Coding augmentation:\\
		$\quad \textbf{for}\quad i=1\;to\;N$\\
		$\quad\quad \textbf{for}\quad j=1\;to\;s_1$ \\
		$\quad\quad\quad {{\bm{\gamma}_i^1({\bm{v}_j})} \leftarrow\left[ {\bm{\gamma} _i^1({\bm{v}_j}),\;\bm{\gamma} _i^1({\bm{v}_j})[\bm{\gamma} _j^{2}({\bm{u}_1}),\cdots ,\bm{\gamma} _j^{2}({\bm{u}_{{s_2}}})]} \right]^\mathsf{T}}$ \\
		$\quad\quad \textbf{end}$ \\
		$\quad \textbf{end}$
	\end{algorithmic}
\end{algorithm}

\section{Experiments}

\begin{figure*}[!h]
	\begin{centering}
		\centering
		\setcounter{subfigure}{0}
		\subfigure{\includegraphics[width=1in]{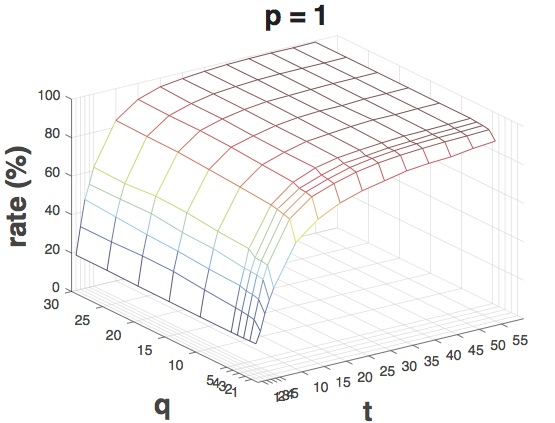}}
		\subfigure{\includegraphics[width=1in]{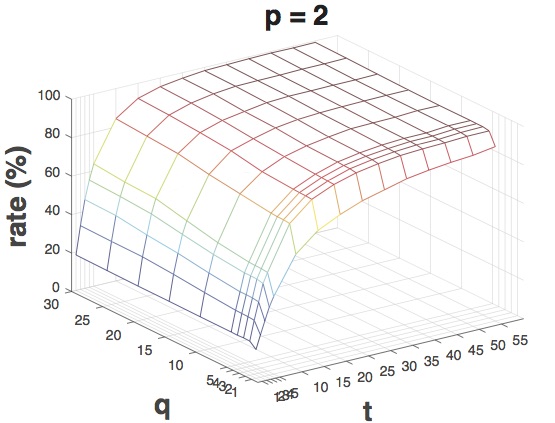}}
		\subfigure{\includegraphics[width=1in]{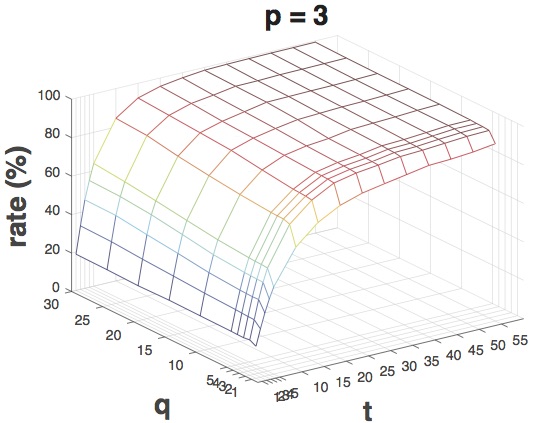}}
		\subfigure{\includegraphics[width=1in]{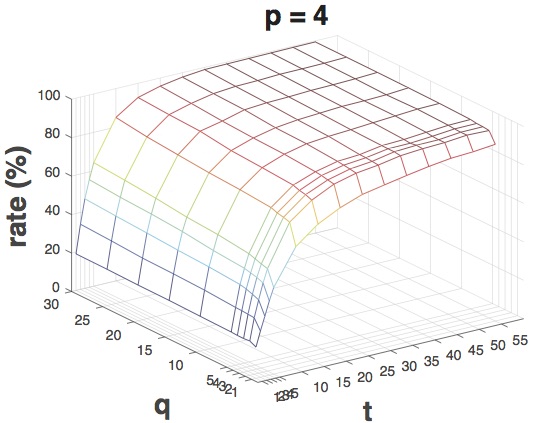}}
		\subfigure{\includegraphics[width=1in]{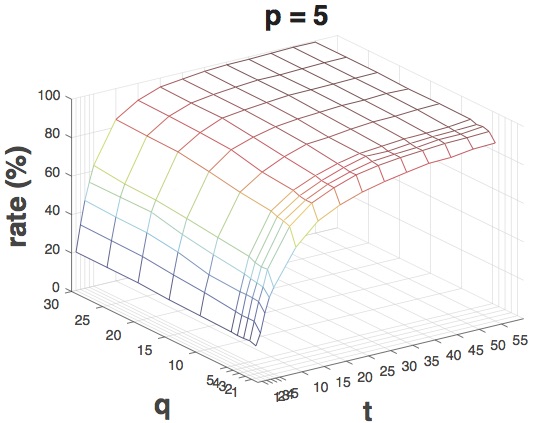}}
		\subfigure{\includegraphics[width=1in]{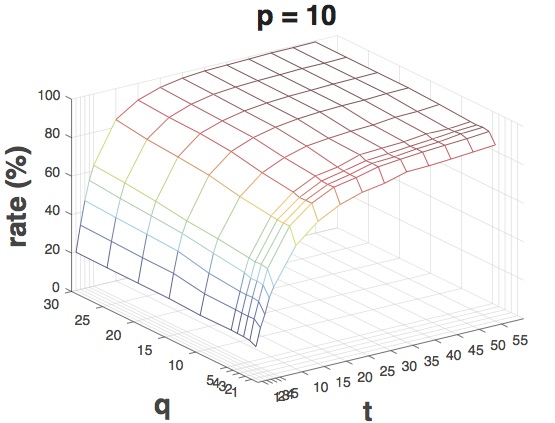}}
		\subfigure{\includegraphics[width=1in]{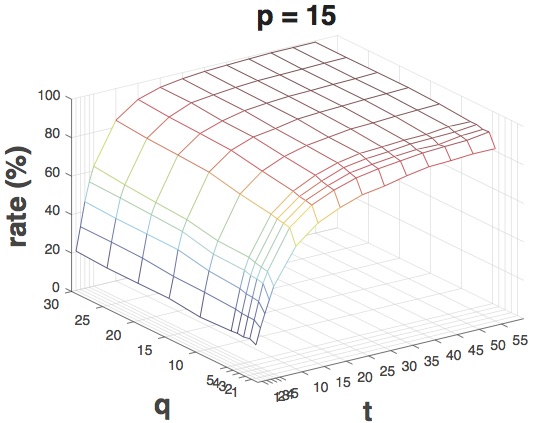}}
		\subfigure{\includegraphics[width=1in]{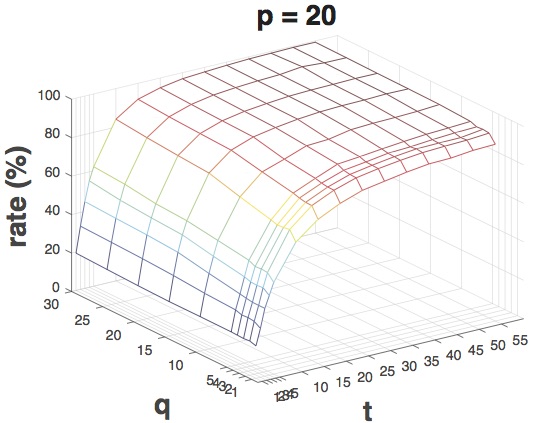}}
		\subfigure{\includegraphics[width=1in]{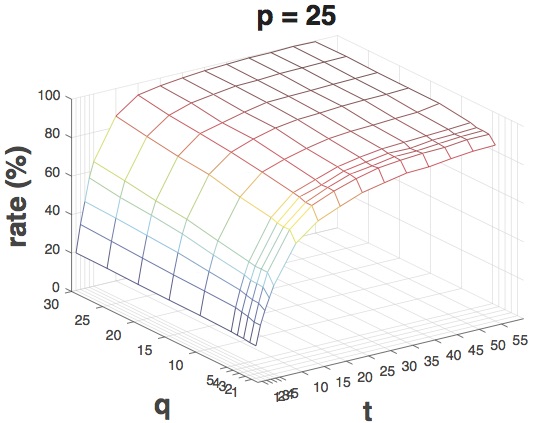}}
		\subfigure{\includegraphics[width=1in]{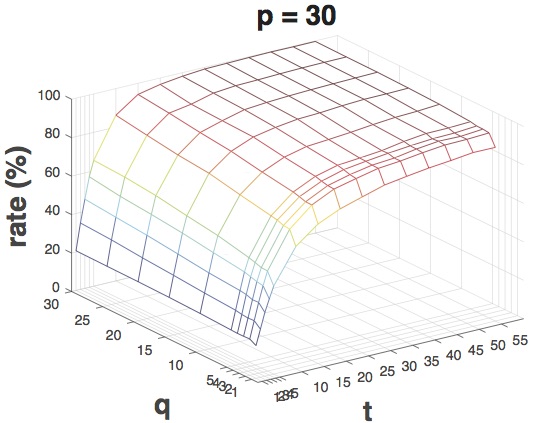}}
		\subfigure{\includegraphics[width=1in]{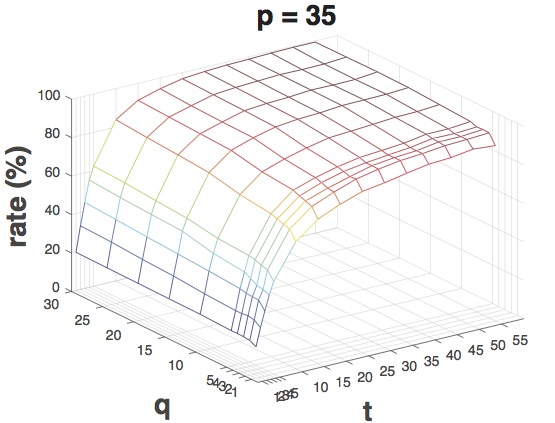}}
		\subfigure{\includegraphics[width=1in]{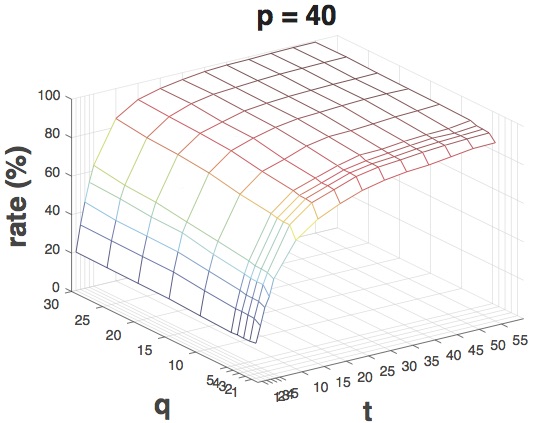}}
		\subfigure{\includegraphics[width=1in]{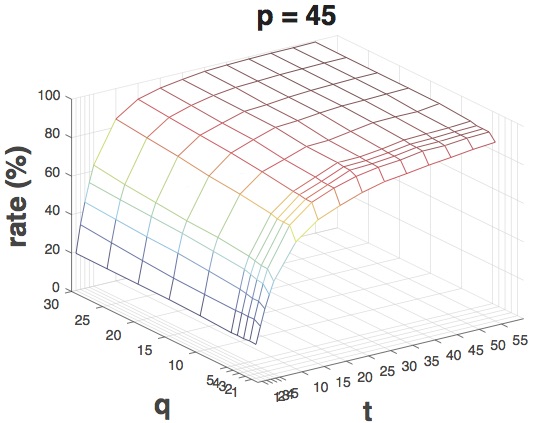}}
		\subfigure{\includegraphics[width=1in]{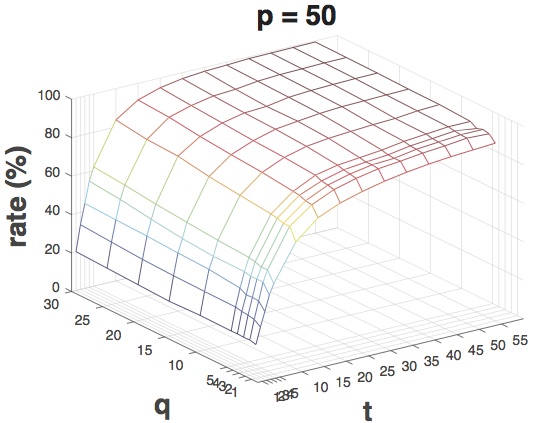}}
		\subfigure{\includegraphics[width=1in]{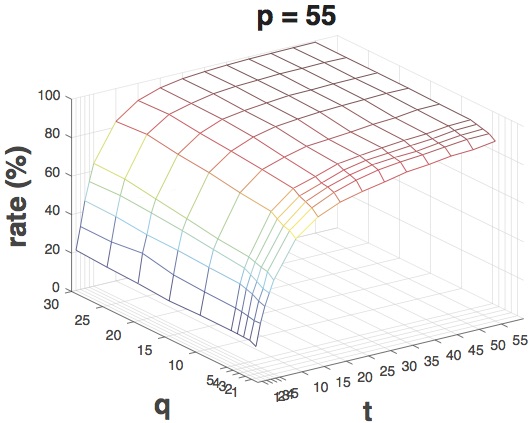}}
		
		\subfigure{\includegraphics[width=1in]{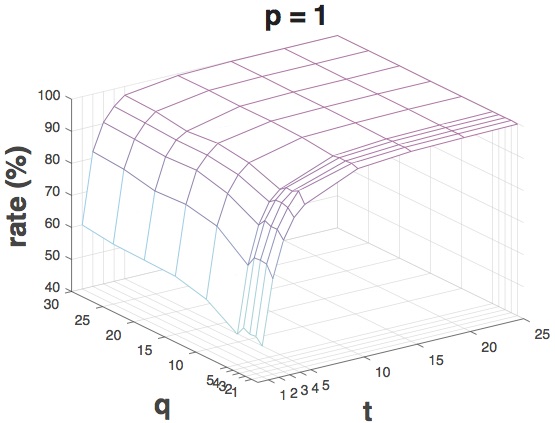}}
		\subfigure{\includegraphics[width=1in]{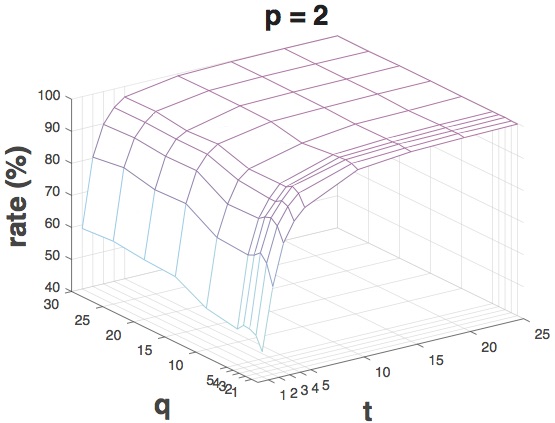}}
		\subfigure{\includegraphics[width=1in]{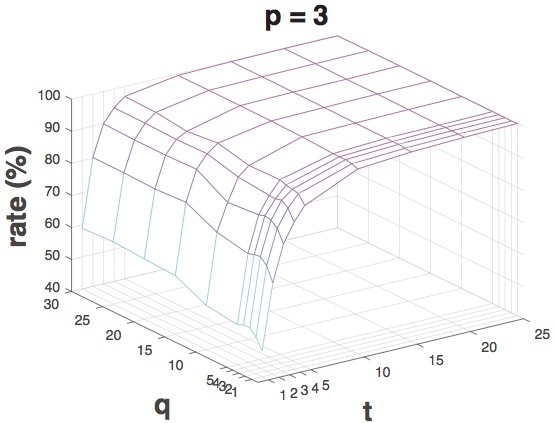}}
		\subfigure{\includegraphics[width=1in]{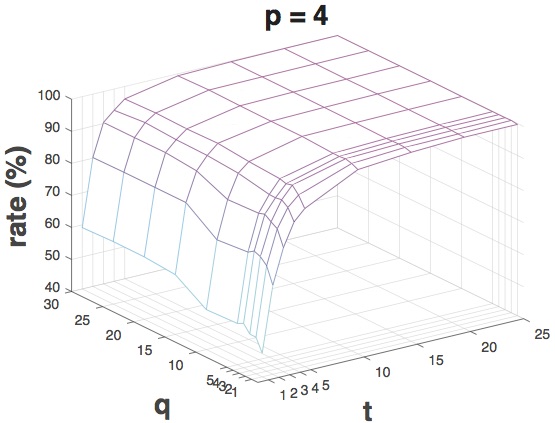}}
		\subfigure{\includegraphics[width=1in]{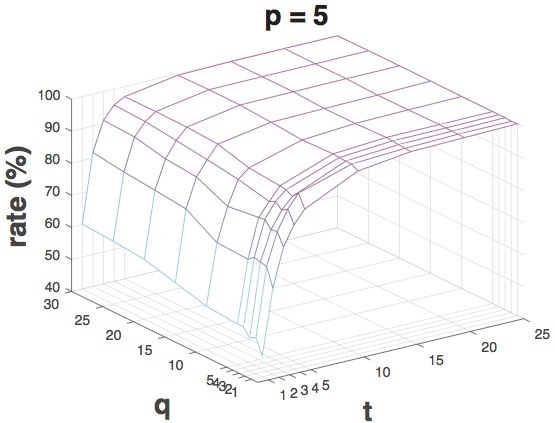}}
		\subfigure{\includegraphics[width=1in]{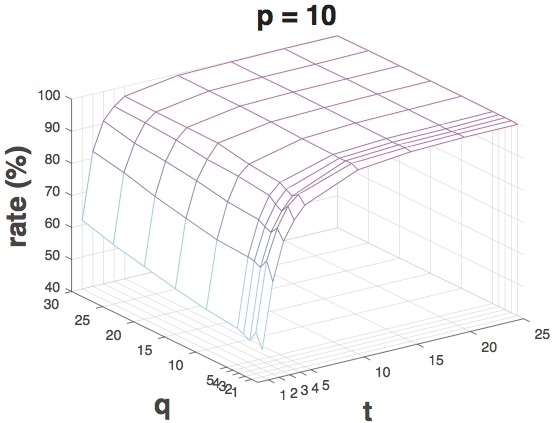}}
		\subfigure{\includegraphics[width=1in]{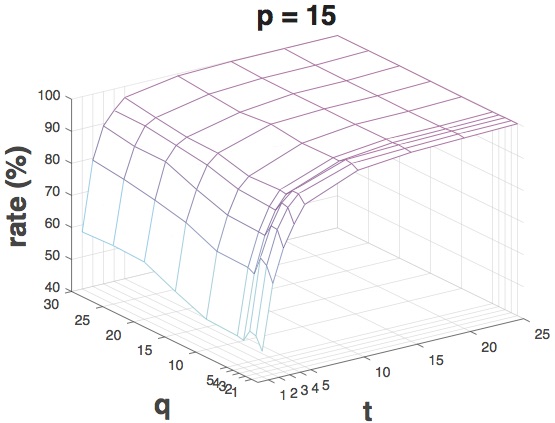}}
		\subfigure{\includegraphics[width=1in]{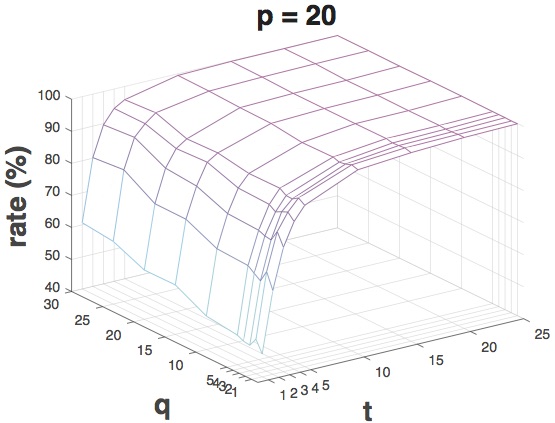}}
		\subfigure{\includegraphics[width=1in]{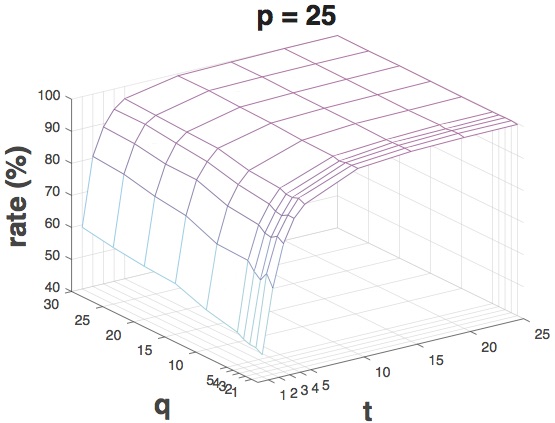}}
		\subfigure{\includegraphics[width=1in]{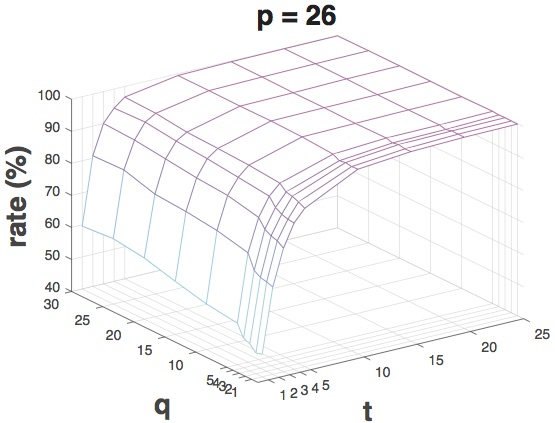}}
		\caption{Classification rate with different parameter $p$ on the Extended YaleB (Top)  and AR Face (Bottom) datasets.}
		\label{Fig:yaleb_ar_triandictsamp}
	\end{centering}
	\vspace{-0.4cm}
\end{figure*}

\subsection{Datasets}
According to \cite{bao2016dictionary,sun2016learning,jian2016semi,akhtarjoint,zhang2017jointly,hu2018nonlinear,xiao2015two}, we evaluate the performance of DDLCN on four popular datasets, which are all standard datasets for dictionary learning evaluation. 

(i) Extended YaleB dataset \cite{georghiades2001few} contains 2,414 frontal face images of 38 people.
There are 59 to 64 images for each person.
All the experiments randomly select half of the images per category as training data and the other half for testing;
(ii) AR Face dataset \cite{MARTINEZ.M.:1998} comprises of over 4,000 color images of 126 people (70 men and 56 women), including frontal views of faces with different facial expressions, lighting conditions and occlusions.
Each person has 26 face images taken during two sessions, in each of which, each person has 13 images.
Among them, 3 are obscured by scarves, 6 by sunglasses, and the remaining faces are of different facial expressions or illumination variations which we refer to as unobscured images.
Following the standard evaluation procedure, we use a subset of the dataset which consists of 2,600 images from 50 male subjects and 50 female subjects.
For each subject, we randomly select 20 samples for training and the other 6 images for testing;
(iii) Caltech 256 dataset \cite{Griffin07tech} contains 30,607 images of 257 categories.
The number of images per category varies from 80 to 827;
(iv) MNIST  dataset \cite{lecun1998gradient} consists of 60,000 training digits and 10,000 testing digits.

\subsection{Parameter  Setting}

In the proposed DDLCN model there are only three parameters 
while CNN-based methods have more parameters that need to be tuned. 
We have conducted exhaustive experiments to emphasize the superiority of the proposed method.
Three parameters for the experiments are, 
\begin{itemize}[leftmargin=*]
	\item $p$, the number of training dictionary samples per category;
	\item $q$, the size of the first layer dictionary per category;
	\item $t$, the number of training samples per category.
\end{itemize}
For simplicity, $p$-$q$ denotes $p$ images are randomly selected per category for training dictionary and $q$ dictionary bases are learned per category in the first dictionary. 
For instance,``15-15'' means $p{=}15$ and $q{=}15$.
The size of the first dictionary is $D_1 {=} q*r$, while the second one is fixed to $D_2 {=} 64$.
The number of the nearest atoms is fixed to 15 and 10 for the first and second feature coding layers, respectively.
We repeat all the experiments 10 times with different random splits of the training and testing images to obtain reliable results.
The final classification rates are reported as the average of each run.

\subsection{Results on Parameter $\bm{p}$}
The first dictionary is trained on $p{=}[1,2,3,4,5,10,15,20,25,30,35,40,45,50,55]$ and $p{=}[1,2,3,4,5,10,15,20,25,26]$ samples per category on the Extended YaleB and AR Face datasets, respectively.
We consider that in this exhaustive way, the superiority of our method could be fully reflected.
From the results shown in Figure~\ref{Fig:yaleb_ar_triandictsamp}, we can clearly see the relationships between q \& p, q \& t and p \& t.
In addition, we consistently observe that the classification accuracy achieves a peak with 10 training samples and then tends to be stable.

\subsection{Results on Parameter $\bm{q}$}
\begin{figure*}[!t]
	\begin{centering}
		\centering
		\setcounter{subfigure}{0}
		\subfigure{\includegraphics[width=1in]{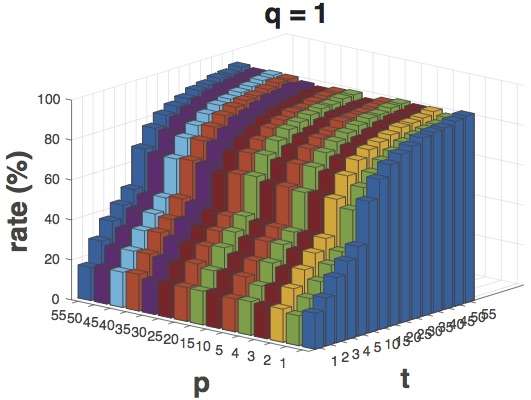}}
		\subfigure{\includegraphics[width=1in]{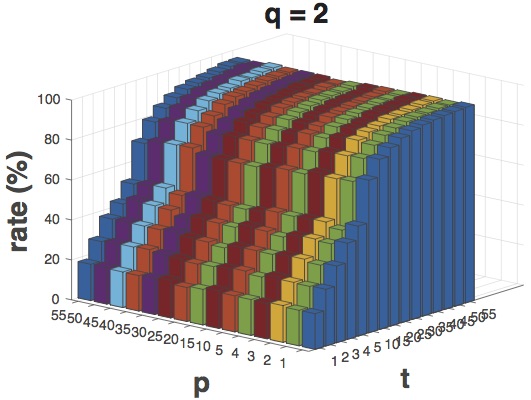}}
		\subfigure{\includegraphics[width=1in]{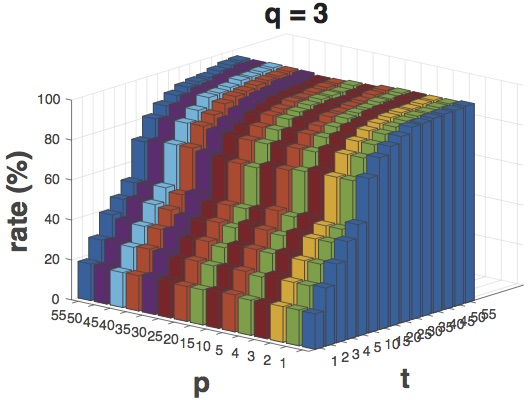}}
		\subfigure{\includegraphics[width=1in]{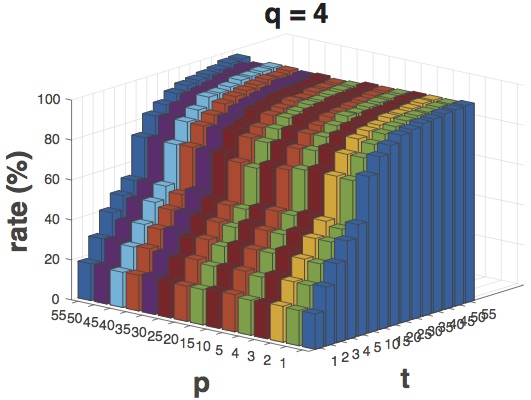}}
		\subfigure{\includegraphics[width=1in]{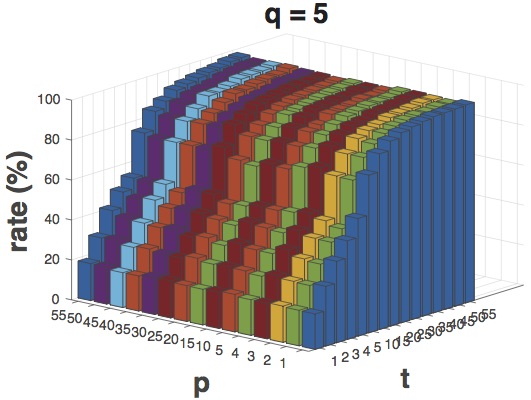}}
		\subfigure{\includegraphics[width=1in]{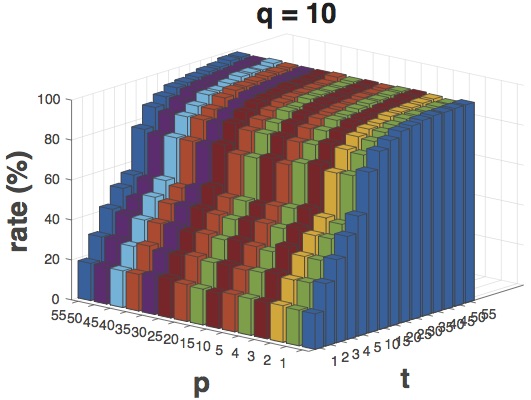}}
		\subfigure{\includegraphics[width=1in]{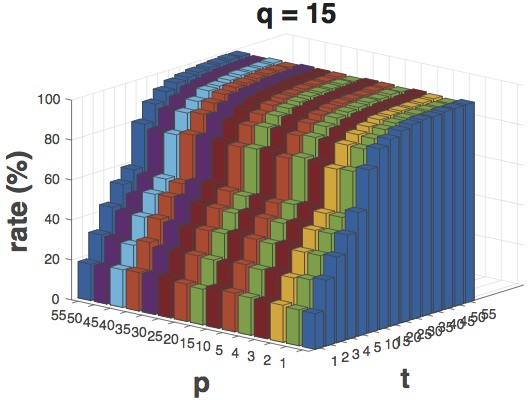}}
		\subfigure{\includegraphics[width=1in]{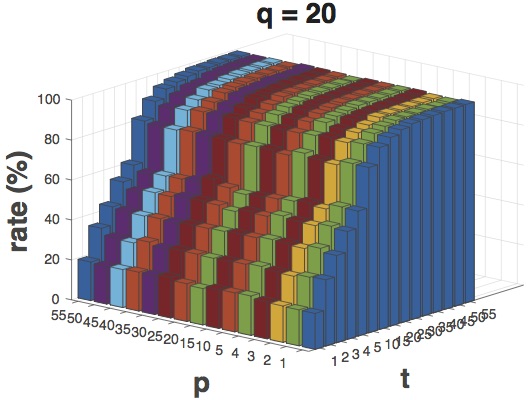}}
		\subfigure{\includegraphics[width=1in]{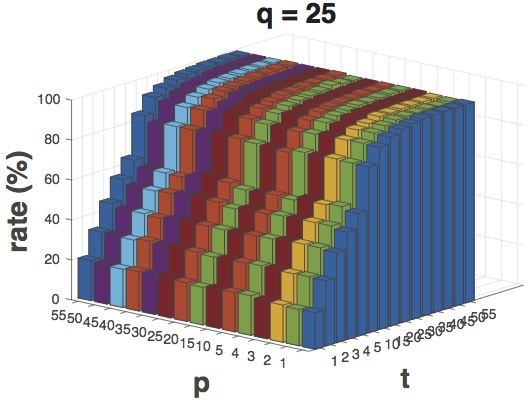}}
		\subfigure{\includegraphics[width=1in]{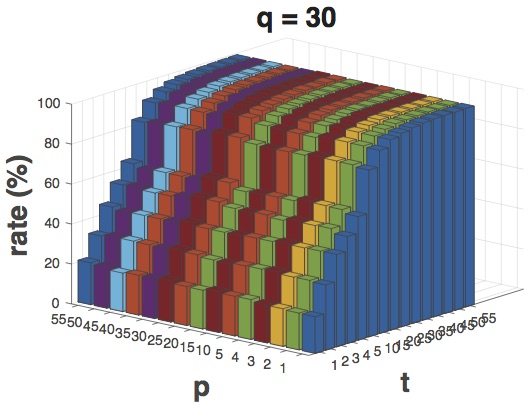}}
		
		\subfigure{\includegraphics[width=1in]{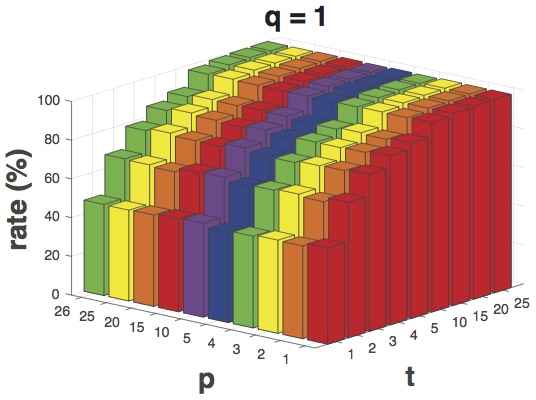}}
		\subfigure{\includegraphics[width=1in]{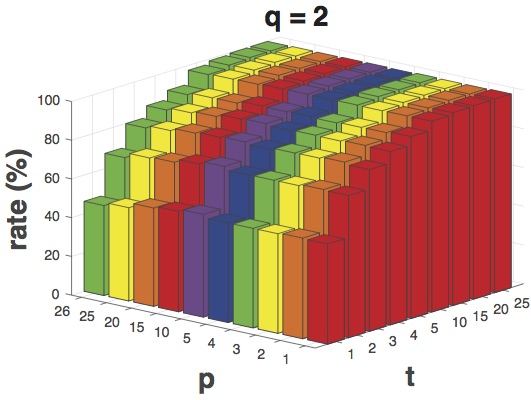}}
		\subfigure{\includegraphics[width=1in]{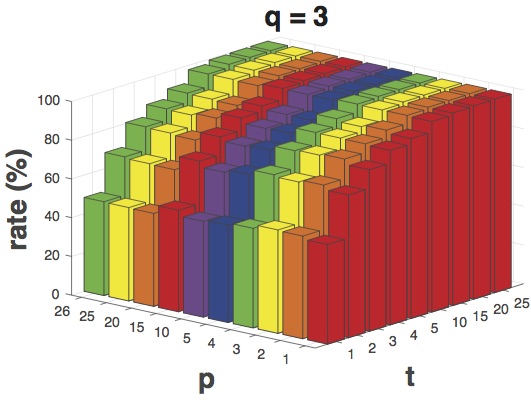}}
		\subfigure{\includegraphics[width=1in]{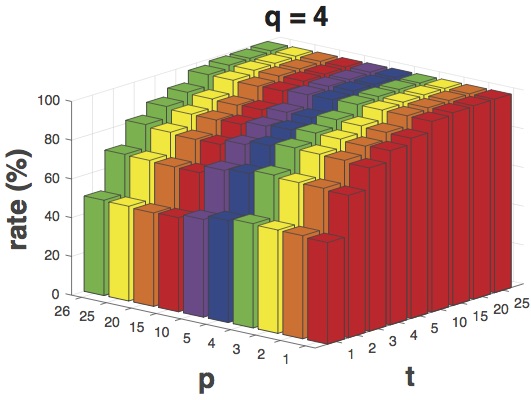}}
		\subfigure{\includegraphics[width=1in]{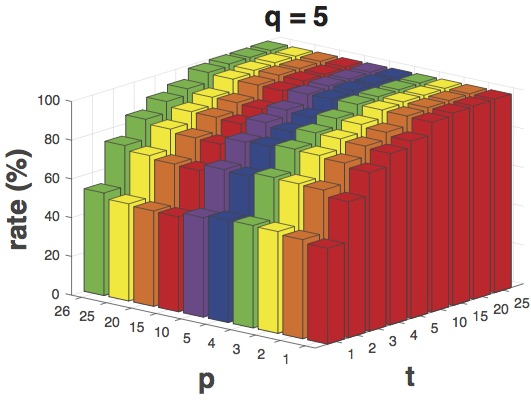}}
		\subfigure{\includegraphics[width=1in]{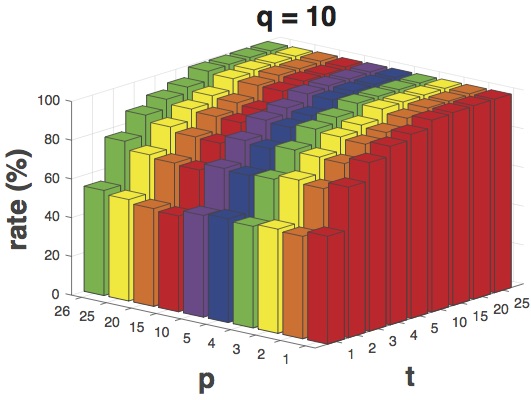}}
		\subfigure{\includegraphics[width=1in]{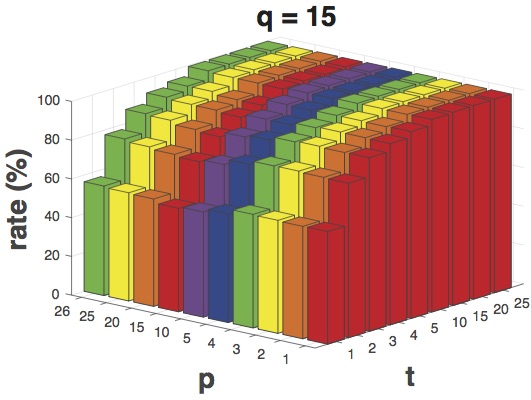}}
		\subfigure{\includegraphics[width=1in]{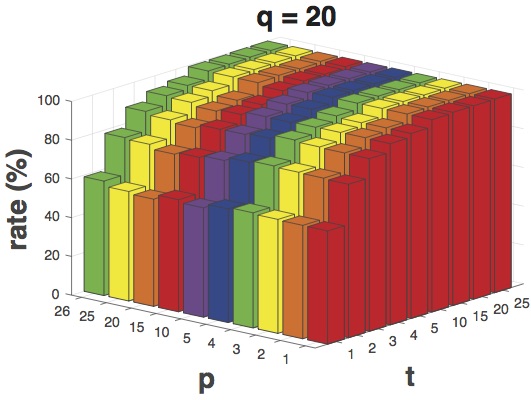}}
		\subfigure{\includegraphics[width=1in]{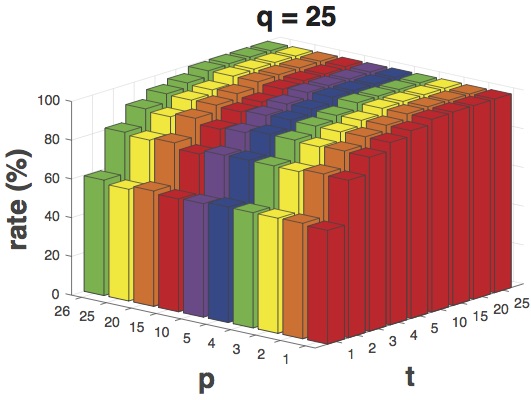}}
		\subfigure{\includegraphics[width=1in]{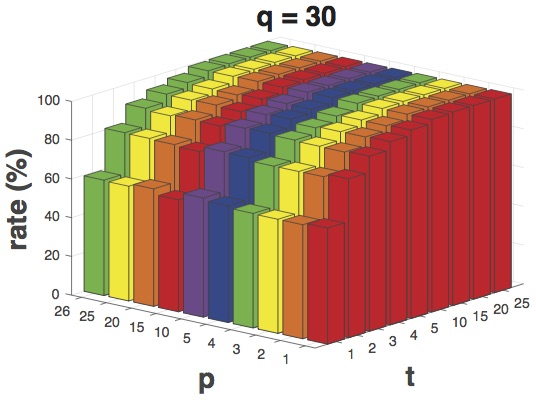}}	
		\caption{Classification rate with varying $q$ on the Extended YaleB (Top) and AR Face (Bottom) datasets.}
		\label{Fig:yaleb_ar_firstdictsize}
	\end{centering}
	\vspace{-0.4cm}
\end{figure*}

We evaluate our approach with different $q {=} [1,2,3,4,5,10,15,20,25,30]$ per person on both the Extended YaleB and AR Face datasets.
The results are shown in Figure~\ref{Fig:yaleb_ar_firstdictsize}.
We observer that on the Extended YaleB dataset that with different $q$, the gaps among all the classification rates are marginal due to the introduction of deep dictionary learning and coding strategy.
This strategy can exploit more information about signal $\bm{y'}$ and incorporate more gradient information about $\bm{y'}$ into the coding structure.
Interestingly, when 20 images per class are randomly selected as training data, the classification rate is close to 100\% when only using 1 atom per person on the AR Face dataset.
The basic reason for the excellent recognition performance is that the proposed DDLCN fully exploits the intrinsic structure of the manifold where features reside, and incorporates more information about the nonlinear function on each group of basis vectors.

\begin{figure*}[!h]
	\begin{centering}
		\centering
		\setcounter{subfigure}{0}
		\subfigure{\includegraphics[width=1in]{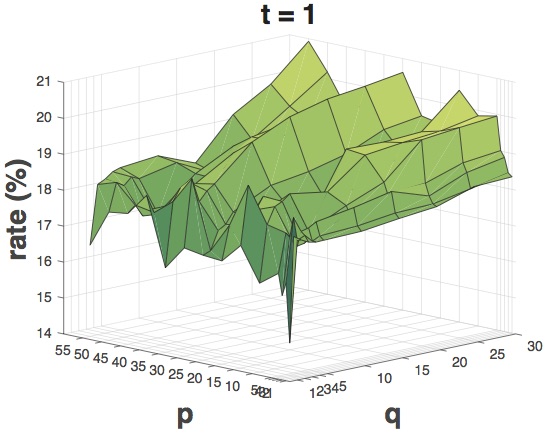}}
		\subfigure{\includegraphics[width=1in]{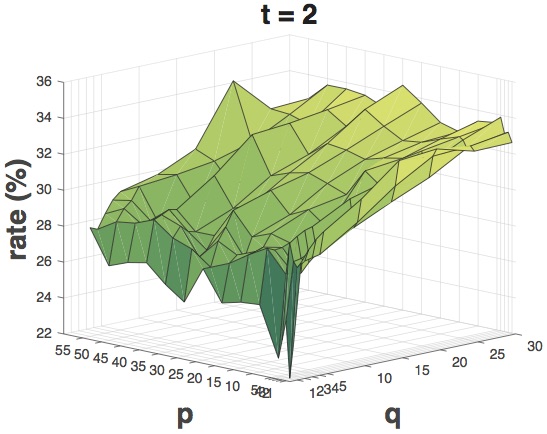}}
		\subfigure{\includegraphics[width=1in]{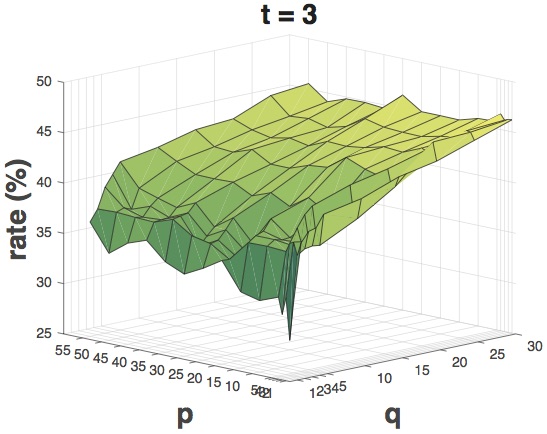}}
		\subfigure{\includegraphics[width=1in]{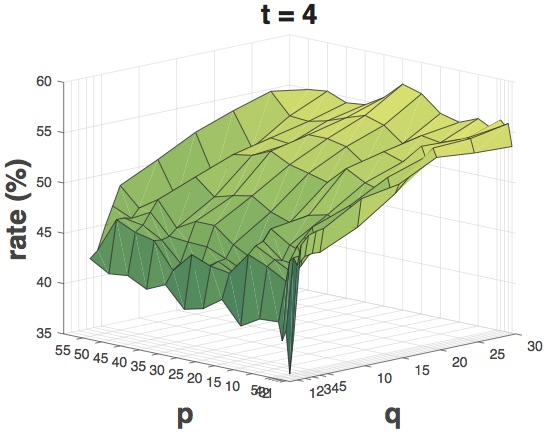}}
		\subfigure{\includegraphics[width=1in]{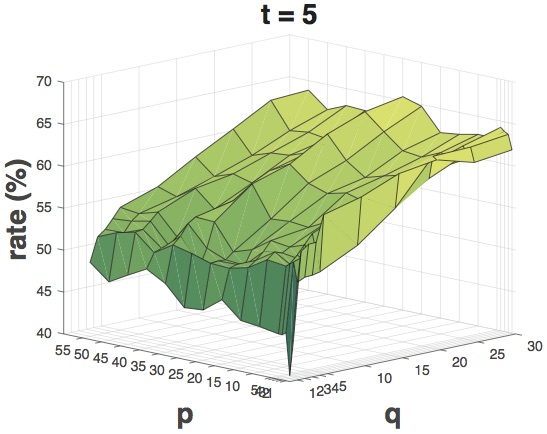}}
		\subfigure{\includegraphics[width=1in]{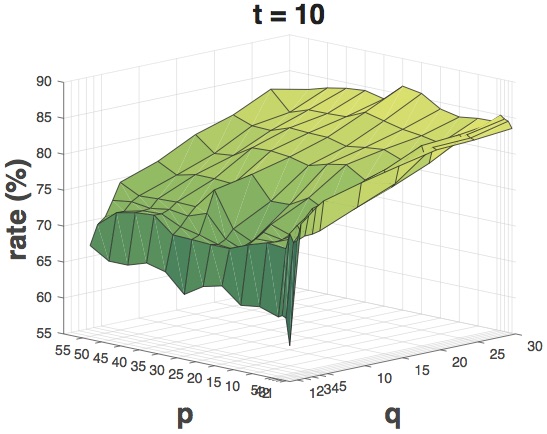}}
		\subfigure{\includegraphics[width=1in]{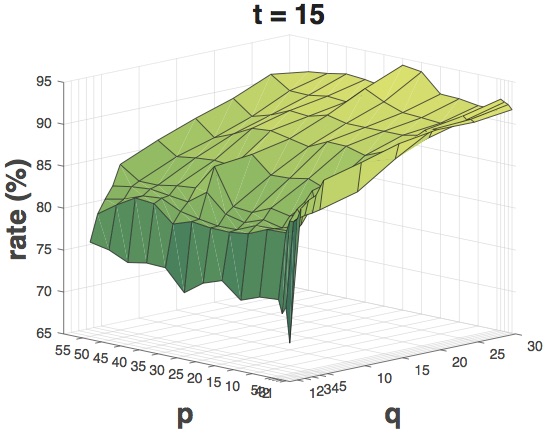}}
		\subfigure{\includegraphics[width=1in]{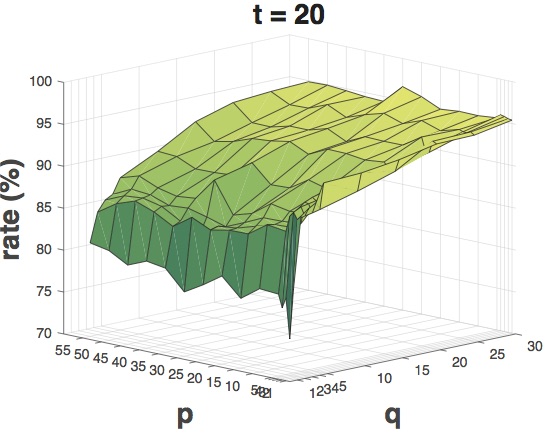}}
		\subfigure{\includegraphics[width=1in]{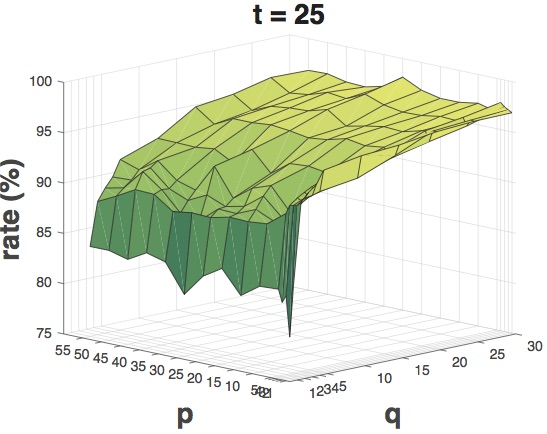}}
		\subfigure{\includegraphics[width=1in]{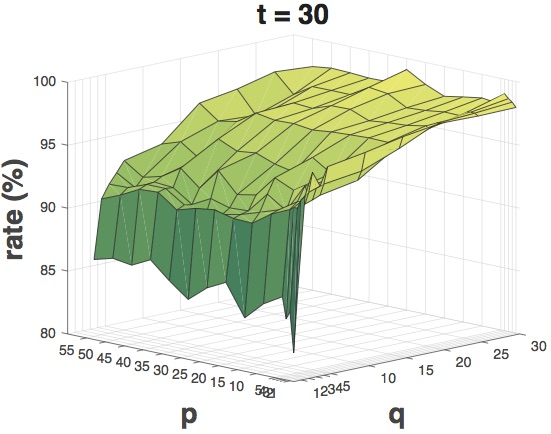}}
		\subfigure{\includegraphics[width=1in]{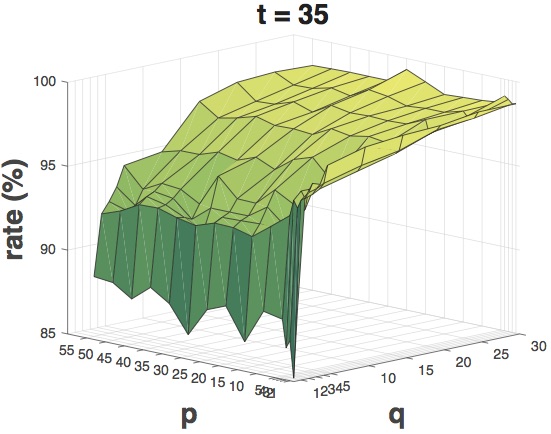}}
		\subfigure{\includegraphics[width=1in]{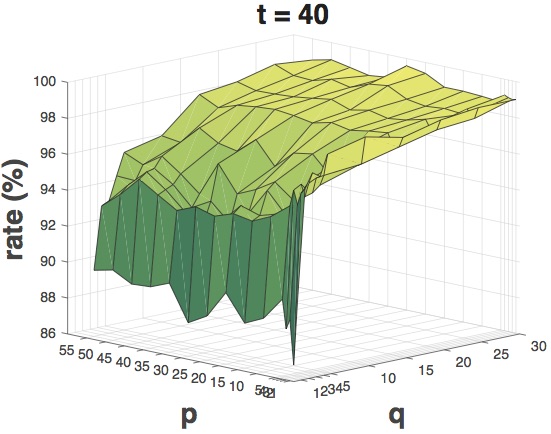}}
		\subfigure{\includegraphics[width=1in]{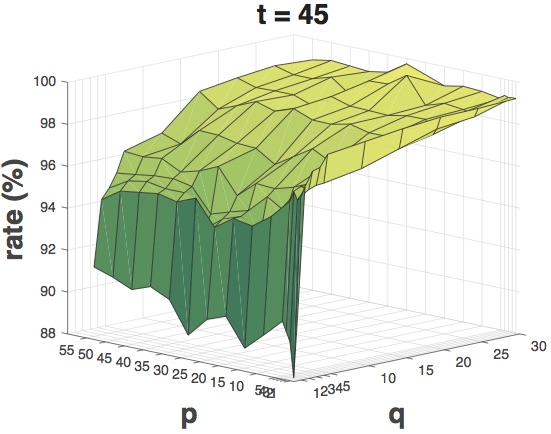}}
		\subfigure{\includegraphics[width=1in]{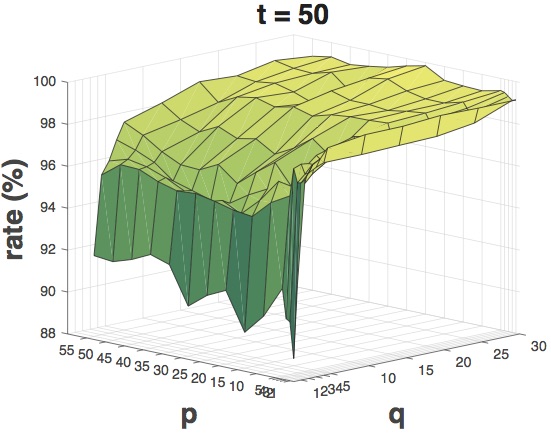}}
		\subfigure{\includegraphics[width=1in]{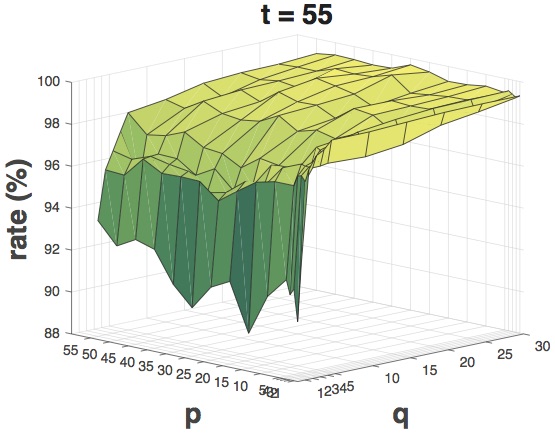}}
		
		\subfigure{\includegraphics[width=1in]{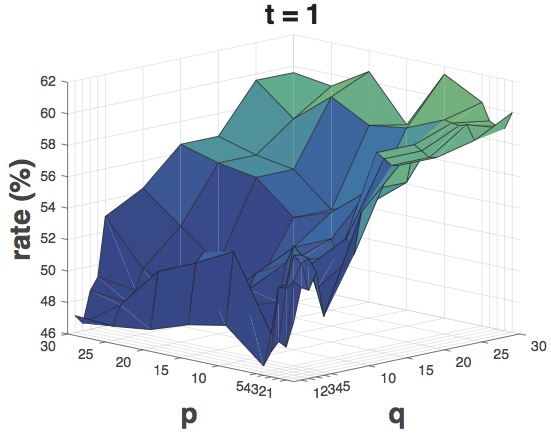}}
		\subfigure{\includegraphics[width=1in]{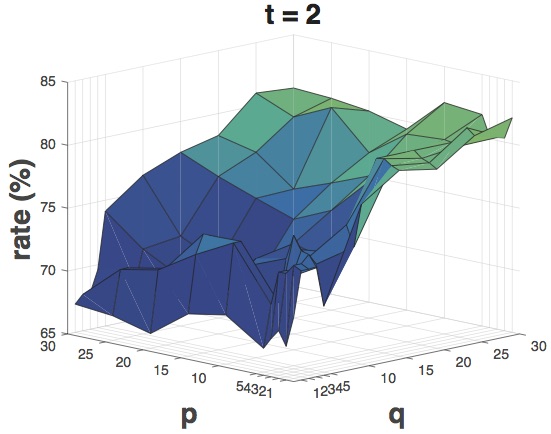}}
		\subfigure{\includegraphics[width=1in]{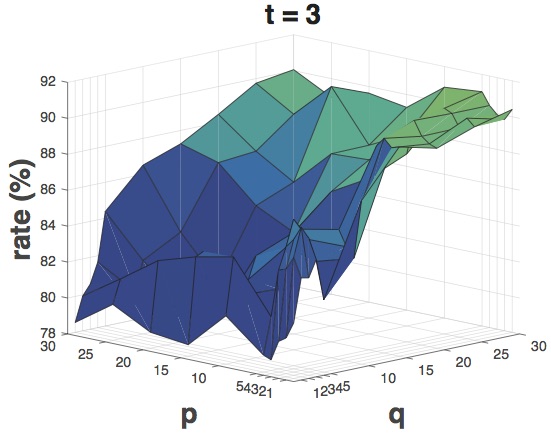}}
		\subfigure{\includegraphics[width=1in]{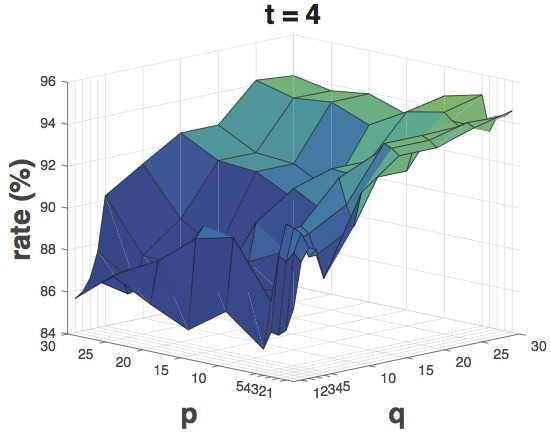}}
		\subfigure{\includegraphics[width=1in]{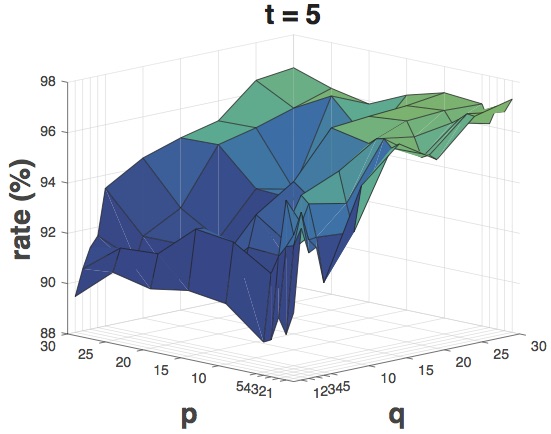}}
		\subfigure{\includegraphics[width=1in]{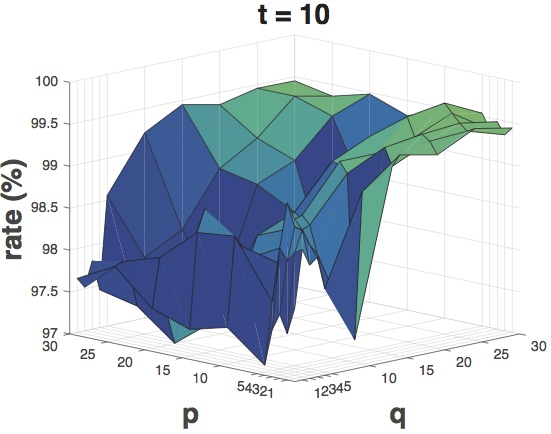}}
		\subfigure{\includegraphics[width=1in]{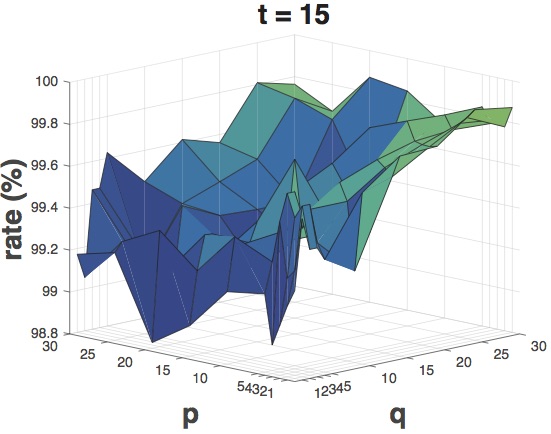}}
		\subfigure{\includegraphics[width=1in]{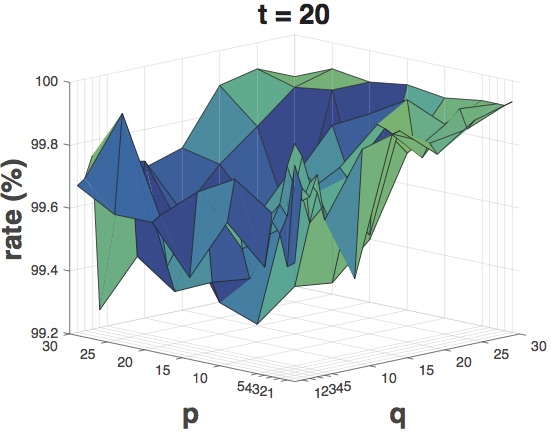}}
		\subfigure{\includegraphics[width=1in]{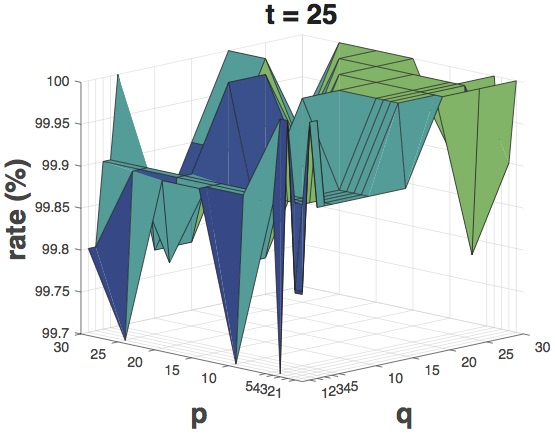}}
		\caption{Classification rate with different $t$ on the Extended YaleB (Top) and AR Face (Bottom) datasets.}
		\label{Fig:yaleb_ar_trainsamp}
	\end{centering}
\end{figure*}

\subsection{Results on Parameter $\bm{t}$}
The number of the training images is $t {=} [1,2,3,4,5,10,15,20,25,30,35,40,45,50,55]$ and $t {=} [1,2,3,4,5,10,15,20,25]$ per category on the Extended YaleB and AR Face datasets, we can draw two conclusions from Figure~\ref{Fig:yaleb_ar_trainsamp},
(i) the classification rate first rises to the peak rapidly and then tends to be stable as $t$ increasing;
(ii) there is a small impact to classification accuracy when changing $p$.

\begin{table}[!t] \scriptsize
	\centering
	\caption{Classification rate (\%) on the Extended YaleB dataset.}
	\resizebox{\linewidth}{!}{%
		\begin{tabular}[!hb]{l|c|c|c}
			\hline  Method                                          & Included (\%)     & Excluded$^{\ast}$ (\%)     & Time (ms) \\ \hline
			SRC (15 per person) \cite{wright2009robust}             & 80.50             & 86.70                      & 11.22 \\
			LLC (30 local bases) \cite{wang2010locality}            & 82.20             & 92.10                      & -    \\
			DL-COPAR \cite{wang2014classification}                  & 86.47 $\pm$ 0.69  & -                          & 31.11 \\
			FDDL \cite{yang2014sparse}                              & 90.01 $\pm$ 0.69  & -                          & 42.48 \\
			LLC (70 local bases) \cite{wang2010locality}            & 90.70             & 96.70                      & -    \\
			DBDL \cite{akhtar2016discriminative}                    & 91.09 $\pm$ 0.59  & -                          & 1.07 \\
			JBDC \cite{akhtarjoint}                                 & 92.14 $\pm$ 0.52  & -                          & 1.02 \\
			K-SVD (15 per person) \cite{aharon2006k}                & 93.10             & 98.00                      & - \\
			SupGraphDL-L \cite{yankelevsky2017structure}            & 93.44             & -                          & - \\	
			D-KSVD (15 per person) \cite{zhang2010discriminative}   & 94.10             & 98.00                      & - \\
			LC-KSVD1 (15-15) \cite{jiang2011learning}               & 94.50             & 98.30                      & 0.52 \\
			LC-KSVD2 (15-15) \cite{jiang2011learning}               & 95.00             & 98.80                      & 0.49 \\
			Multi-Block Alternating Proximal\cite{bao2016dictionary}& 95.12             & -                          & - \\ 
			ITDL \cite{QiangQiu2014}                                & 95.39             & -                          & - \\ 
			VAE + GAN \cite{mathieu2016disentangling}               & 96.4              & -                          & - \\
			EasyDL \cite{quan2016sparse}                            & 96.22             & -                          & - \\	
			LC-KSVD2 (A-15) \cite{jiang2011learning}                & 96.70             & 99.00                      & - \\	
			SRC (all training samples) \cite{wright2009robust}      & 97.20             & 99.00                      & 20.78 \\
			CSDL-SRC(power) \cite{liu2016face}                      & 98.28 $\pm$ 0.57  & -                          & - \\
			RRC\_L$_1$ (300) \cite{yang2013regularized}             & 99.80             & -                          & - \\ \hline
			PCANet-1 \cite{chan2015pcanet}                          & 97.77             & -                          & - \\
			PCANet-2 \cite{chan2015pcanet}                          & \textbf{99.85}    & -                          & - \\ \hline
			DDLCN (1-1)                                             & 87.42 $\pm$ 1.33  & 89.54 $\pm$ 1.02           & \textbf{0.18} \\
			DDLCN (15-15)                                           & 97.38 $\pm$ 0.54  & 98.48 $\pm$ 0.48           & 0.71 \\
			DDLCN (55-15)                                           & 97.68 $\pm$ 0.60  & 98.64 $\pm$ 0.52           & 0.92 \\
			DDLCN (A-15)                                            & 98.34 $\pm$ 0.56  & \textbf{99.18 $\pm$ 0.46}  & 0.98 \\ \hline
	\end{tabular}}
	\label{table:extendedYaleB}
	\vspace{-0.3cm}
\end{table}

\subsection{Comparison Against Baselines}
We compare our results with the state-of-the-art, the comparative results of the Extended YaleB, AR Face, Caltech 256 and MNIST datasets.

\noindent \textbf{Extended YaleB.} 
We compare our DDLCN with the traditional dictionary learning methods, \emph{e.g.}, D-KSVD \cite{zhang2010discriminative}, LC-KSVD \cite{jiang2011learning}, and other deep learning approaches, \emph{e.g.}, PCANet \cite{chan2015pcanet} and VAE + GAN \cite{mathieu2016disentangling}.
DDLCN is better than all the baselines except \cite{chan2015pcanet} and \cite{yang2013regularized} as shown in the second column of Table~\ref{table:extendedYaleB}.
A different training strategy is adopted which helps improve the classification performance in \cite{chan2015pcanet}.
Even under this unequal conditions, our method also outperforms the PCANet-1 (97.77\%) and is slightly worse than PACNet-2 (99.85\%), which validates the advantages of our method.
Followed the evaluation metric of \cite{jiang2011learning}, another experiment with the bad images excluded (10 for each person) is performed and the results are listed in the third column of Table~\ref{table:extendedYaleB}.
It is observed that the proposed DDLCN achieves higher classification rate than other methods when using the A-15 strategy.
In addition, we compare with SRC  \cite{wright2009robust}, LC-KSVD \cite{jiang2011learning}, DBDL \cite{akhtar2016discriminative}and
JBDC \cite{akhtarjoint} in terms of the computation time for classifying one test image, as shown in the fourth column of Table \ref{table:extendedYaleB}.
The time that our approach took is much less than LC-KSVD, DBDL, JBDC and other methods.

\noindent \textbf{AR Face.} 
We compare the DDLCN with some advanced methods, \emph{e.g.,} LC-KSVD \cite{jiang2011learning}, SupGraphDL-L \cite{yankelevsky2017structure}, \emph{etc}.  
We can observe that our approach outperforms others including \cite{chan2015pcanet} and \cite{yang2013regularized} when only using 1-1 strategy in Table~\ref{table_AR}.
It is a surprising result and the reason is that the proposed DDLCN makes the approximation quality from one layer $O({\varepsilon ^2})$, ascending to two layers $O({\varepsilon ^3})$ according to Lipschitz Smoothness.
This ensures the quality of approximation to achieve $O({\varepsilon ^3})$ on imperfect atoms.
In addition, we also report the computation time (ms) for classification on several methods.
As shown in the third column of Table~\ref{table_AR}, the time that our approach took is marginally more than LC-KSVD, but is much less than SRC, DL-COPAR, JBDC, DBDL and FDDL.

\begin{table}[!t] \scriptsize
	\centering
	\caption{Classification rate (\%) on the AR Face dataset.}
	\resizebox{\linewidth}{!}{%
		\begin{tabular}{l|c|c}
			\hline  Method                                            & Accuracy (\%)             & Time (ms) \\ \hline
			SRC (5 per person) \cite{wright2009robust}                & 66.50                     & 17.76 \\			
			LLC (30 local bases) \cite{wang2010locality}              & 69.50                     & - \\
			DL-COPAR \cite{wang2014classification}                    & 83.29 $\pm$ 1.23          & 36.49 \\
			FDDL \cite{yang2014sparse}                                & 85.97 $\pm$ 1.23          & 50.03 \\	
			DBDL \cite{akhtar2016discriminative}                      & 86.15 $\pm$ 1.19          & 1.20 \\
			K-SVD (5 per person) \cite{aharon2006k}                   & 86.50                     & - \\
			JBDC \cite{akhtarjoint}                                   & 87.17 $\pm$ 0.99          & 1.18 \\
			LLC (70 local bases) \cite{wang2010locality}              & 88.70                     & - \\
			D-KSVD (5 per person) \cite{zhang2010discriminative}      & 88.80                     & - \\
			LC-KSVD1 (5-5) \cite{jiang2011learning}                   & 92.50                     & 0.541 \\
			LC-KSVD2 (5-5) \cite{jiang2011learning}                   & 93.70                     & \textbf{0.479} \\
			Multi-Block Alternating Proximal\cite{bao2016dictionary}  & 93.88                     & - \\ 
			RRC\_L$_1$\cite{yang2013regularized}                      & 96.30                     & - \\				
			ADDL (5 items, 20 labels) \cite{zhang2017jointly}         & 97.00                     & - \\
			SRC (all training samples) \cite{wright2009robust}        & 97.50                     & 83.79 \\
			LC-KSVD2 (A-5) \cite{jiang2011learning}                   & 97.80                     & - \\
			LGII \cite{Soodeh15}                                      & 99.00                     & - \\ \hline
			PCANet-1 \cite{chan2015pcanet}                            & 98.00                     & - \\
			PCANet-2 \cite{chan2015pcanet}                            & 99.50                     & - \\ \hline
			DDLCN (1-1)                                               & 99.56 $\pm$ 0.21          & 0.73 \\
			DDLCN (5-5)                                               & 99.84 $\pm$ 0.36          & 1.26 \\
			DDLCN (A-5)                                               & \textbf{99.87 $\pm$ 0.19} & 1.63 \\ \hline
	\end{tabular}}
	\label{table_AR}
	\vspace{-0.2cm}
\end{table}

\begin{table}[!t]\scriptsize
	\centering
	\caption{Classification rate (\%) on the Caltech 256 dataset.}
	\resizebox{\linewidth}{!}{%
		\begin{tabular}{l|c|c|c|c}
			\hline  Num. of train. samp.            & 15 train         & 30 train         & 45 train         & 60 train \\ \hline
			
			KC \cite{van2008kernel}                 & -                & 27.17 $\pm$ 0.46 & -                & - \\
			LLC \cite{wang2010locality}             & 25.61            & 30.43            & -                & - \\
			K-SVD \cite{aharon2006k}                & 25.33            & 30.62            & -                & - \\
			D-KSVD \cite{zhang2010discriminative}   & 27.79            & 32.67            & -                & - \\
			LC-KSVD1 \cite{jiang2011learning}       & 28.10            & 32.95            & -                & - \\
			SRC \cite{wright2009robust}             & 27.86            & 33.33            & -                & - \\			
			Griffin \cite{griffin2007caltech}       & 28.30            & 34.10 $\pm$ 0.20 & -                & - \\
			LC-KSVD2 \cite{jiang2011learning}       & 28.90            & 34.32            & -                & - \\
			Graph-matching \cite{duchenne2011graph} & -                & 38.10 $\pm$ 0.60 & -                & -  \\
			Local NBNN \cite{mccann2012local}       & 33.50 $\pm$ 0.90 & 40.10 $\pm$ 0.10 & -                & - \\ 			
			Latent Structural\cite{liu2016describing}& 36.34           & 42.01            & -                & -  \\		
			ScSPM \cite{yang2009linear}             & 27.73 $\pm$ 0.51 & 34.02 $\pm$ 0.35 & 37.46 $\pm$ 0.55 & 40.14 $\pm$ 0.91 \\
			NDL \cite{hu2018nonlinear}              & 29.30 $\pm$ 0.29 & 36.80 $\pm$ 0.45 & -                & - \\
			LScSPM \cite{gao2010local}              & 30.00 $\pm$ 0.14 & 35.74 $\pm$ 0.10 & 38.54 $\pm$ 0.36 & 40.43 $\pm$ 0.38 \\
			SSC \cite{Gabriel14}                    & 30.60 $\pm$ 0.30 & 37.00 $\pm$ 0.30 & 40.70 $\pm$ 0.10 & 43.50 $\pm$ 0.30 \\
			SNDL \cite{hu2018nonlinear}             & 31.10 $\pm$ 0.35 & 38.25 $\pm$ 0.43 & -                & - \\
			MLCW \cite{fanello2014ask}              & 34.10            & 39.90            & 42.40            & 45.60 \\
			CRBM \cite{sohn2011efficient}           & 35.1             & 42.1             & 45.7             & 47.9 \\
			LP-$\beta$\cite{gehler2009feature}      & -                & 45.8             & -                & - \\	
			M-HMP \cite{bo2013multipath}            & 42.7             & 50.7             & 54.8             & 58.0 \\ \hline
			Convolutional Networks \cite{zeiler2014visualizing}    & - & -                & -                & 74.2 $\pm$ 0.3 \\
			VGG19  \cite{simon2015neural}           & -                & -                & -                & \textbf{84.10} \\ \hline
			
			DDLCN (1-1)   & 26.30 $\pm$ 0.40             & 31.45 $\pm$ 0.21           & 34.69 $\pm$ 0.31          & 37.76 $\pm$ 0.25        \\
			DDLCN (15-15) & 35.06 $\pm$ 0.26  & 41.26 $\pm$ 0.22  & 44.17 $\pm$ 0.35 & 47.48 $\pm$ 0.26 \\
			DDLCN (30-30) & \textbf{45.25 $\pm$ 0.31}  & \textbf{51.64 $\pm$ 0.51}  & \textbf{55.11 $\pm$ 0.26} & 59.66 $\pm$ 0.45 \\ \hline
	\end{tabular}}
	\label{table_Caltech256}
	\vspace{-0.2cm}
\end{table}

\begin{figure}[!t]
	\centering
	\centering
	\setcounter{subfigure}{0}
	\subfigure[car side,   acc: 100.00\%]{\includegraphics[width=0.48\linewidth]{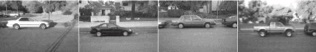}}
	\subfigure[faces easy, acc: 99.20\%]{\includegraphics[width=0.48\linewidth]{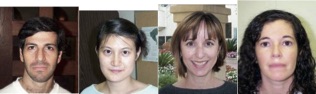}}\\
	\subfigure[leopards,   acc: 98.46\%]{\includegraphics[width=0.48\linewidth]{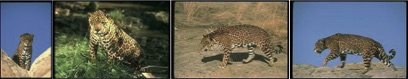}}
	\subfigure[motorbikes, acc: 98.29\%]{\includegraphics[width=0.48\linewidth]{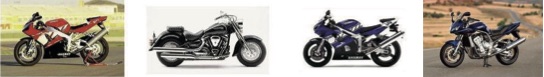}}\\
	\subfigure[airplanes,  acc: 98.03\%]{\includegraphics[width=0.48\linewidth]{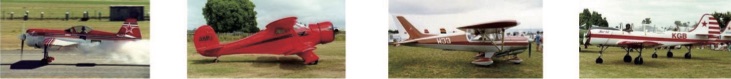}}
	\subfigure[tower pisa,     acc: 97.33\%]{\includegraphics[width=0.48\linewidth]{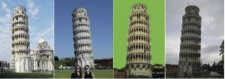}}\\
	\subfigure[ketch,      acc: 97.16\%]{\includegraphics[width=0.48\linewidth]{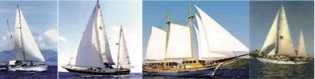}}
	\subfigure[brain,      acc: 96.30\%]{\includegraphics[width=0.48\linewidth]{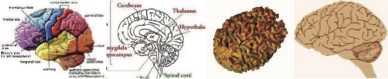}}\\
	\subfigure[sunflower,  acc: 96.00\%]{\includegraphics[width=0.48\linewidth]{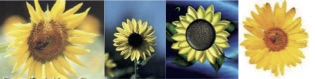}}
	\subfigure[watch,      acc: 95.36\%]{\includegraphics[width=0.48\linewidth]{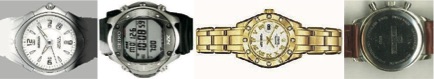}}\\
	\caption{Example images from classes with high classification accuracy from the Caltech 256 dataset.}
	\label{Fig: caltech256_top10}
	\vspace{-0.3cm}
\end{figure}

\noindent \textbf{Caltech 256.}
We evaluate our approaches on 15, 30, 45 and 60 training images per class and compare with the state of the art, the comparisons are shown in Table~\ref{table_Caltech256}.
The proposed method achieves better results than other traditional dictionary learning methods in all cases except VGG Net \cite{simon2015neural} and convolutional network \cite{zeiler2014visualizing} when using 60 training samples.
In \cite{simon2015neural}, the authors use a very deep convolutional network (up to 19 layers) for the task, which reveals that the network depth is of crucial importance, and the leading results on the challenging dataset all exploit ``very deep'' models.
Beside, note that after extracting features, the feature learner and coder of the DDLCN would be a fixed block that would not be updated during training, and only the linear classifier on top is updated during training. 
Compared with \cite{simon2015neural} and \cite{zeiler2014visualizing}, both are CNN-based models and they learn features directly from the raw pixels.
Both methods need to train the weights for the entire network rather than only the weights of linear classifier on the top layer, being as such computationally expensive.
However, the training of the DDLCN is offline which represents a big advantage. 
The test phase of the DDLCN is pretty fast for our method. 
Moreover, our approaches outperform the other competing dictionary learning approaches even when using the 15-15 strategy, including K-SVD, D-KSVD, LC-KSVD, LLC, etc.
What is surprising is that even when using the 1-1 strategy, our method still achieves better results than K-SVD, KC and CRBM, which validates the advantages of the proposed method.
Figure~\ref{Fig: caltech256_top10} shows some example images from classes with high classification accuracy.

\begin{table}[!t] \tiny
	\centering
	\caption{Classification rate (\%) on the MNIST dataset.}
	\resizebox{0.76\linewidth}{!}{%
		\begin{tabular}{l|c} \hline
			Method                                                    & Accuracy (\%)  \\ \hline
			Deep Representation Learning \cite{yang2015deep}          & 85.47        \\
			RGF \cite{MingyuFan2014A}                                 & 98.09          \\ 
			DCN \cite{lin2010deep}                                    & 98.15          \\ \hline
			Embed CNN \cite{weston2012deep}                           & 98.50          \\
			Convolutional Clustering \cite{dundar2015convolutional}   & 98.60 \\
			
			Deep Convolutional Learning \cite{pu2015generative}       & \textbf{99.58} \\ \hline
			DDLCN (1-2)       & 96.56  \\
			DDLCN (100-100)   & 98.55  \\
			DDLCN (500-500)   & 99.02  \\ \hline
	\end{tabular}}
	\vspace{-0.4cm}
	\label{table_mnist}
\end{table}

\begin{figure}[!t]
	\centering
	\centering
	\setcounter{subfigure}{0}
	\subfigure{\includegraphics[width=0.49\linewidth]{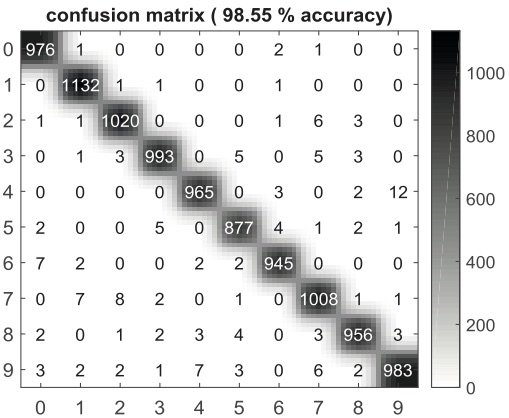}}
	\subfigure{\includegraphics[width=0.49\linewidth]{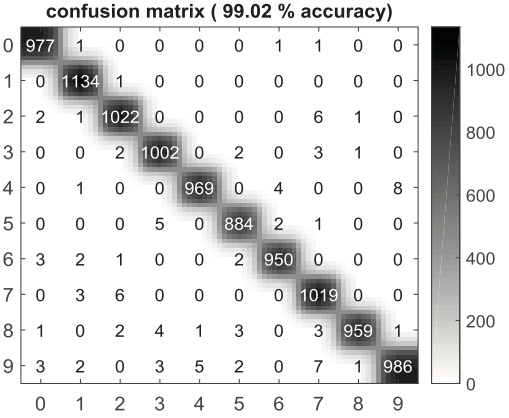}}\\
	\caption{Confusion matrix on the MNIST dataset.}
	\label{Fig:mnist_con_500_500}
	\vspace{-0.5cm}
\end{figure}

\noindent \textbf{MNIST.} The classification rates of different approaches on the MNIST dataset are provided in Table~\ref{table_mnist}.
We observe that the DDLCN consistently outperforms all the baselines except \cite{dundar2015convolutional} and \cite{pu2015generative} when using the100-100 strategy.
The authors in \cite{pu2015generative} use a two layers model and plus a one layer features to achieve better result (+1.03\%) than us.
This is because both the CNN-based methods are jointly optimized between forward and backward propagation, and the proposed method has no end-to-end tuning.
Thus the training of DDLCN is more efficieint than CNN-based methods.
Such deep networks integrate low/mid/highlevel features and classifiers under supervision.
In addition, when $p{=}1$ and $q{=}2$, we even achieve 96.56\% classification accuracy, which proves again that our method can also achieve good recognition rate when the number of the training samples is limited and the size of the dictionary is small.
This phenomenon will be of great benefit in practical applications, especially when the training data are limited.
Note that when $p{=}100$ and $q{=}100$, the classification rate is boosted to 98.55\% (Figure \ref{Fig:mnist_con_500_500} (a)).
When $p{=}500$ and $q{=}500$, the classification rate is further improved to 99.02\%.
Figure \ref{Fig:mnist_con_500_500} (b) shows the confusion matrix and we observe that the most confused pairs are (2, 7), (4, 9) and (3, 5).

\section{Conclusion}
The goal of this paper is to improve the deep representation capability of  dictionary learning. 
To this end, we propose a novel deep dictionary learning network DDLCN to learn multi-layer deep dictionaries, which combines the advantages of both deep learning and dictionary learning and achieves impressive performance. 
We designed a dictionary learning layer and used it to replace traditional convolutional layers in a CNN.
More specifically, for two-layer framework, the first layer learns a dictionary to represent input data, the next layer learns a dictionary to represent the atoms of the first dictionary. 
For a given input sample, the code consists of its locality constrained sparse code, together with the locality constrained sparse codes for all of the atoms that participate in its code. 
Experimental results on four popular benchmarks demonstrate that the proposed DDLCN outperforms the existing dictionary learning methods and achieves competitive results compared with the CNN-based models.
Our code is available at \url{https://github.com/Ha0Tang/DDLCN}.

%\clearpage
{\small
\bibliographystyle{ieee}
\bibliography{egbib}
}

\end{document}